\documentclass[journal]{IEEEtran}

\usepackage{times}  
\usepackage{helvet}  
\usepackage{courier}  
\usepackage{url}  
\usepackage{graphicx}  
\usepackage{bm}
\usepackage{amsmath,amssymb,amsfonts}
\usepackage{multirow}
\usepackage{multicol}
\usepackage{enumitem}

\usepackage{bbm}
\usepackage{color}
\usepackage{microtype}
\newcommand{\tabincell}[2]{\begin{tabular}{@{}#1@{}}#2\end{tabular}}

\ifCLASSINFOpdf
\else
\fi
\begin{document}
		
		\newcommand{\cname}{\emph{R$^2$-Net}}
		\newcommand{\fcname}{\emph{Relation of Relation Learning Network}}
		
		\newcommand{\jname}{\emph{DELE}}
		\newcommand{\fjname}{\emph{\textbf{D}escription-\textbf{E}nhanced \textbf{L}abel \textbf{E}mbedding network}}
		%
		\title{Description-Enhanced Label Embedding Contrastive Learning for Text Classification}
		%
		%
		%
		
		\author{Kun~Zhang,~\IEEEmembership{Member,~IEEE},
			Le~Wu,~\IEEEmembership{Member,~IEEE},  
			Guangyi~Lv, 
			Enhong~Chen,~\IEEEmembership{Senior~Member,~IEEE},
			Shulan~Ruan,
			Jing Liu,~\IEEEmembership{Member,~IEEE},
			Zhiqiang Zhang, Jun Zhou, and
			Meng Wang,~\IEEEmembership{Fellow,~IEEE}
			\IEEEcompsocitemizethanks{\IEEEcompsocthanksitem Kun Zhang, Le Wu, and Meng Wang are with School of Computer and Information, Hefei University of Technology, Hefei, Anhui 230029, China. (email: zhang1028kun, lewu.ustc, eric.mengwang@gmail.com).
				\IEEEcompsocthanksitem Guangyi Lv is with AI Lab at Lenovo Research, Beijing, 100094, China. (email: lvgy1@lenovo.com)
				\IEEEcompsocthanksitem Enhong Chen and Shulan Ruan are with the School of Computer Science and Technology, University of Science and Technology of China, Hefei 230026, China. (email: slruan@mail.ustc.edu.cn, cheneh@ustc.edu.cn).
				\IEEEcompsocthanksitem Jing Liu is with National Laboratory of Pattern Recognition, Institute of Automation, Chinese Academy of Sciences, Beijing, 100190, China. (email: jliu@nlpr.ia.ac.cn)
				\IEEEcompsocthanksitem Zhiqiang Zhang and Jun Zhang are with Ant Group CO., Ltd, Hangzhou, 310007, China (email: lingyao.zzq, jun.zhoujun@antfin.com)
				
				Le Wu is the corresponding author.
			}
		}
		
		\markboth{Journal of \LaTeX\ Class Files,~Vol.~14, No.~8, January~2021}%
		{Shell \MakeLowercase{\textit{et al.}}: Bare Demo of IEEEtran.cls for IEEE Journals}
		%



		\maketitle
		\begin{abstract}
			
			Text Classification is one of the fundamental tasks in natural language processing, which requires an agent to determine the most appropriate category for input sentences.
			Recently, deep neural networks have achieved impressive performance in this area, especially Pre-trained Language Models (PLMs).
			Usually, these methods concentrate on input sentences and corresponding semantic embedding generation. 
			However, for another essential component: labels, most existing works either treat them as meaningless one-hot vectors or use vanilla embedding methods to learn label representations along with model training, underestimating the semantic information and guidance that these labels reveal.
			To alleviate this problem and better exploit label information, in this paper, we employ Self-Supervised Learning~(SSL) in model learning process and design a novel self-supervised \textbf{R}elation of \textbf{R}elation~(R$^2$) classification task for label utilization from a one-hot manner perspective. 
			Then, we propose a novel \fcname~(\cname) for text classification, in which text classification and R$^2$ classification are treated as optimization targets. 
			Meanwhile, triplet loss is employed to enhance the analysis of differences and connections among labels. 
			Moreover, considering that one-hot usage is still short of exploiting label information, we incorporate external knowledge from WordNet to obtain multi-aspect descriptions for label semantic learning and extend \cname~to a novel \fjname~(\jname) from a label embedding perspective. 
			One step further, since these fine-grained descriptions may introduce unexpected noise, 
			we develop a mutual interaction module to select appropriate parts from input sentences and labels simultaneously based on Contrastive Learning~(CL) for noise mitigation. 
			Extensive experiments on different text classification tasks reveal that \cname~can effectively improve classification performance, and \jname~can make better use of label information and further improve the performance. 
			As a byproduct, we have released the codes\footnote{https://github.com/little1tow/DELE\_pytorch} to facilitate other research.
		\end{abstract}
		

		%
		\IEEEpeerreviewmaketitle
		
		\section{Introduction}
		\label{s:introduction}
\begin{figure*}
			\centering
			\includegraphics[width=0.77\textwidth]{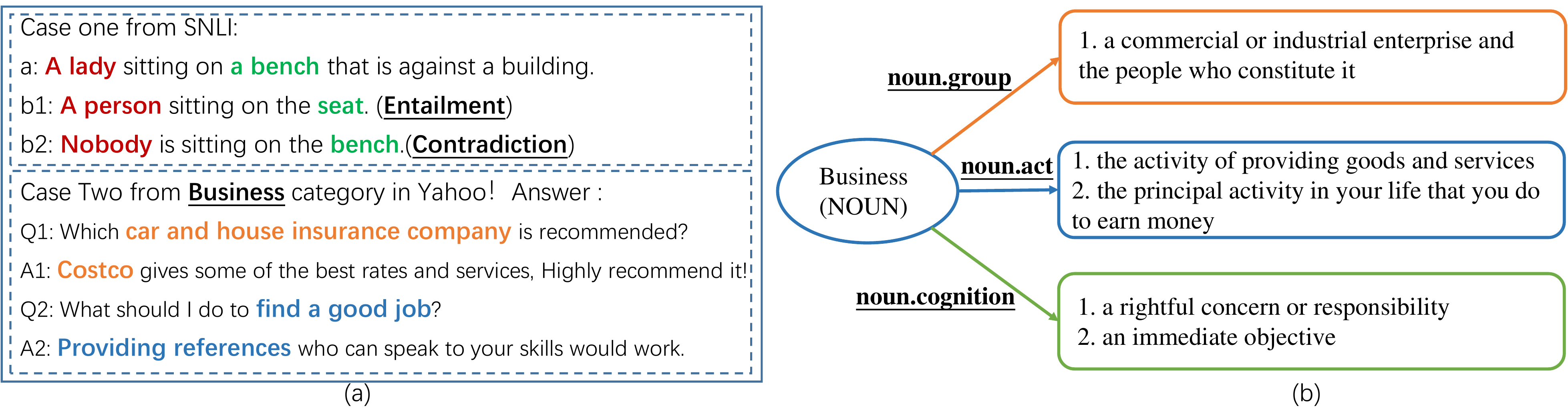}
			\caption{(a) Some text classification examples: label signals can imply specific some semantic expressions of input sentences (e.g., In Entailment pair (\textit{a} and \textit{b1}): ``A lady'' is replaced with ``A person'', In Contradiction pair (\textit{a} and \textit{b2}): ``A lady'' is replaced with ``Nobody'' ).   (b) Some fine-grained noun descriptions for label word ``Business'' from WordNet, which give detailed explanations about different attributes of label ``Business''.}
			\label{f:example}
		\end{figure*}

		As one of fundamental tasks in Natural Language Processing~(NLP), text classification focuses on identifying the most appropriate category for sentences or the most suitable relation for sentence pairs. 
		For example, Paraphrase Identification~(PI) aims at identifying whether the sentence pair expresses the same meaning (Yes or No)~\cite{dolan2005automatically}. 
		Natural Language Inference~(NLI) targets at classifying input sentence pair into one of three relations~(i.e., \textit{Entailment, Contradiction, Neutral})~\cite{Kim2018SemanticSM}. 
		QA Topic Classification requires an agent to select the most suitable topic for a given question-answer pair~\cite{chen2017reading}. 
		Fig.~\ref{f:example}(a) illustrates some examples with different relations from different tasks. 
		
		As a vital technology, text classification has been applied successfully to various NLP fields, e.g., sentiment analysis~\cite{yilmaz2023multi,huang2022attention,zhu2021senti}, information retrieval~\cite{Liu2010LearningTR}, question answering~\cite{liu2018finding}, and dialogue system~\cite{serban2016building}. 
		Based on label usage, most of the existing work can be grouped into two categories.  
		The first category is \textit{one-hot encoding of labels}.
		Researchers usually focus on designing a deep model to learn representations for input text~\cite{zhang2019drr,zhang2021ladra,tan2022dynamic}.
		Then, a simple classifier is employed to predict the label distribution.
		Cross-Entropy loss between the prediction and one-hot label encoding is finally adopted for model training~\cite{zhang2018ImageEnhance,guo2020label}. 
		However, labels can reveal some common characteristics of examples within the same category~\cite{xiong2021fusing} and provide rich semantic information as well as guidance for sentence semantics learning~\cite{zhang-etal-2018-multi}.
		Treating labels as independent and meaningless one-hot vectors will cause potential information loss and have weaknesses in dealing with fine-grained interactions between text and labels. 
		Gururangan \textit{et al.}~\cite{gururangan2018annotation} have observed that different relations among sentence pairs imply specific semantic expressions. 
		Taking Case one in Fig.~\ref{f:example}(a) as an example, 
		when constructing sentence pairs that are semantically contradictory, negation words (e.g., \textit{replacing ``A lady'' with ``Nobody''} in pair \textit{a} and \textit{b2}) are usually used. 
		Moreover, replacing exact numbers with approximates can always generate ``\textit{entailment}'' semantic relation. 
		Therefore, the connection and differences among different relations~(e.g., pairwise relation comparison) will be helpful to capture more implicit common semantic features, which is a promising direction to fully exploit label information for text classification.
		Then, the problem becomes how to design this type of signal and integrate it into model learning process, which is one of the main focuses of this paper. 
		
		The second category is \textit{label embedding method}. 
		Different from one-hot encoding methods, this type of work usually uses dense vectors to represent labels in the same semantic space as text embeddings. 
		Then, attention mechanism is employed to measure the impact of label semantics on input text for classification.
		For example,  Xiao \textit{et al.}~\cite{xiao2019label} proposed a label-specific attention network to enhance text representation learning. 
		Zhang \textit{et al.}~\cite{zhang2022long} proposed to retrieve keywords from documents as pseudo descriptions for label embedding learning. 
		Despite the inspiring performance, there still are some shortcomings. 
		Specifically, vanilla embedding methods treat labels as general words and generate representations during model training. 
		Knowledge-enhanced methods usually use a single description to enrich the embedding. 
		They either only focus on the literal meanings of labels or describe labels in a coarse-grained manner, which is insufficient for precise and relevant label embedding learning. 
		As shown in Fig.~\ref{f:example}(b), even if a particular meaning of a word is utilized as label semantics, this meaning still has different attributes and can be described from different perspectives (e.g., \textit{noun.group}, \textit{noun.act}). 
		Moreover, even if different input texts have the same labels, their semantics may associate with different aspects of the label. 
		For example, for label ``\textit{Business}'', \textit{noun.group} attribute is used to identify QA(1), and \textit{noun.act} attribute is used for QA(2) in case two in Fig.~\ref{f:example}(a). 
		Thus, it is promising to consider fine-grained knowledge~(e.g., multi-aspect descriptions from WordNet) to enhance label embedding.
		Meanwhile, different from key words extraction and single description usage, multi-aspect descriptions may not all relate to the specific label meaning, which will import unanticipated noise, and harm the precise label semantic representation.
		For example, \textit{noun.cognition} attribute in Fig.~\ref{f:example}(b) is not the option for the label ``\textit{Business}'' used in Fig.\ref{f:example}(a), which will become noise and confuse models to understand the semantics of the label ``\textit{Business}''. 
		Therefore, another main contribution of this paper is how to introduce fine-grained information (i.e., multi-aspect descriptions) for better label 
		embedding learning, as well as eliminating unexpected noise. 
		
		In order to mine the implicit information inside labels in a one-hot encoding manner, in our preliminary work~\cite{zhang2021making}, we propose a novel \fcname~(\cname) to fully exploit labels in a simple but effective way. 
		We first utilize PLMs encoder (e.g., BERT~\cite{devlin2018bert}) and CNN-based encoder to model global and partial semantic meanings of input words and sentences separately. 
		Then, inspired by SSL, we propose a self-supervised \textbf{R}elation of \textbf{R}elation~(R$^2$) classification task to enhance the learning ability of \cname, so that inter-class relations of input texts will be modelled comprehensively (i.e., differences among different label relations). 
		Moreover, a triplet loss is used to measure the connections among the same classes~(e.g., those inputs that have the same label will be represented much closer and vice versa further apart).

		However, \cname~still mines label information in a one-hot encoding manner, which inhibits the potential of label utilization. 
		To this end, in this paper, we focus on label embedding and extend \cname~to a novel \fjname~(\jname), which incorporates fine-grained descriptions from WordNet\footnote{https://wordnet.princeton.edu/} as prior knowledge to boost the label embedding learning. 
		Specifically, we first extract multiple descriptions from WordNet for each label word to enrich the label embedding learning. 
		After that, considering that multi-aspect descriptions may not all relate to the specific meaning of labels, we design a novel mutual interaction module based on CL framework to explore bidirectional interactions between input sentences and labels.
		Along this line, sentence representations can be leveraged to denoise multi-aspect descriptions of labels and select the most relevant parts for better semantic representations. 
		Meanwhile, label semantics can also be used to guide the importance selection of input sentences as existing methods do. 
		Extensive experiments over different types of text classification tasks also prove that \jname~can make better use of labels and do better classification.
		
		In summary, the main contributions of this paper lie in the following parts: 
		1) We develop a novel self-supervised R$^2$ task to mine the implicit information inside labels from a one-hot encoding perspective (Preliminary Work); 
		2) We propose to leverage multi-aspect descriptions from WordNet to fully exploit label information from label embedding perspective (Extended Work); 
		3) We also design a novel mutual interaction module based on CL for better description denoising and integrating (Extended Work); 
		4) Extensive experiments over different type of text classification tasks have been done to demonstrate the effectiveness of our proposed methods.
		
		The remainder of this paper is organized as follows. 
		Section \ref{s:related_work} summarizes related work. 
		Section \ref{s:problem} gives formal definitions of text classification and our proposed R$^2$ classification.
		Sections \ref{s:cmodel} and \ref{s:jmodel} report technical details of our proposed \cname~and \jname. 
		The	experiments and detailed analysis are reported in Section \ref{s:experiment}. Finally, we discuss and conclude our work in Sections \ref{s:discussion} and \ref{s:conclusion}.

		\section{Related Work}
		\label{s:related_work}
		In this section, we will introduce the related work, which is grouped into two lines of literature, 1) \textit{Text Classification}: focusing on sentence semantic modelling and relation identification with different label usages; 2) \textit{Contrastive Learning}: introducing the recent progress of CL on text classification.

		\subsection{Text Classification}
		With the development of various neural networks such as CNN~\cite{kim2014convolutional}, GRU~\cite{Chung2014EmpiricalEO}, and attention mechanism~\cite{Parikh2016ADA,huang2022attention}, plenty of methods have been exploited for text classification on large datasets like SNLI~\cite{bowman2015large},  Quora~\cite{iyer2017first}, and SST-5~\cite{socher2013recursive}. 
		Usually, researchers employ neural networks~\cite{devlin2018bert,Liu2019RoBERTaAR} to generate text representations. 
		Then, a simple classifier, such as Multi-Layer Perceptron~(MLP), is used to predict the label distribution. 
		Next, a cross-entropy loss is applied for model training~\cite{zhang2018ImageEnhance,xu2020semi}.
		Based on label usage, there are two categories: \textit{One-hot Encoding methods} and \textit{Label Embedding methods}, which are summarized as follows:
		
		For \textit{One-hot Encoding methods}, 
		researchers focus on input text and treat labels as one-hot meaningless training signals. 
		For example, 
		Zeng \textit{et al.}~\cite{zeng2014relation} focused on the input sentences and generated representations by extracting lexical-level and sentence-level features. 
		Zhang \textit{et al.}~\cite{zhang2019drr} developed a DRr-Net to select important parts in a sentence precisely for text representation and classification. 
		Similarly, Li \textit{et al.}~\cite{liu2020finding} leveraged a sequential decision process with reinforcement learning to exploit the potential of sentences for classification. 
		After PLMs~(e.g., BERT~\cite{devlin2018bert}, RoBERTa~\cite{Liu2019RoBERTaAR}) have been proposed, this kind of learning paradigm has become much simpler and more effective.
		They have all achieved impressive performances.
		However, in most scenarios, treating labels as independent and meaningless one-hot training signals will ignore the implicit semantics and common feature indication of label words. 
		This will cause information loss and limit the learning capability of representation methods.

		To better exploit label information, \textit{label embedding methods} are proposed.
		In practice, researchers treat text classification problem as a label-text joint embedding problem, where labels are embedded in the same space as texts.
		Then, attention mechanism is employed to measure semantic relations between labels and texts for classification~\cite{xiao2019label,cai2020hybrid,mueller2022label,liu2022co,zhu2022generating}.
		Traditionally, vanilla embedding is the most common choice. 
		For example, Du \textit{et al.}~\cite{du2019explicit} used vanilla embedding to represent label semantics from scratch and designed an EXAM to explicitly calculate matching scores between text and labels at word level. 
		To generate better label embedding, side information, as well as relation modelling are taken into consideration. 
		For side information, Zhu \textit{et al.}~\cite{zhu2022generating} used TF-IFD to select the most effective words from documents as the descriptions of labels, and leveraged attention mechanism to fuse the input sentences and label information for better classification. 
		Rivas \textit{et al.}~\cite{rivas2020efficient} extracted textual definitions from Oxford Dictionaries for label modelling. 
		For relations, Guo \textit{et al.}~\cite{guo2020label} developed a label confusion model to estimate the label dependency for better label exploration. 
		In addition, there still exist other methods that leveraged label embedding to tackle multi-label classification~\cite{xiong2021fusing,wu2022effective}, hierarchical text classification~\cite{wang2021concept}, and so on.
		
		Compared with existing work, our proposed work has following main improvements. First, we propose to use multi-aspect real sentences as descriptions for the fine-grained side information of labels. 
		Second, we argued that additional descriptions would import unexpected noise and harm the model performance. 
		Then, we developed a novel mutual interaction based on CL to achieve the fusing of input sentences and labels, as well as alleviate the unexpected noise introduced by multi-aspect descriptions.

		\subsection{Contrastive Learning}
		As a core component of self-supervised learning, CL has led to state-of-the-art performance in the unsupervised training of deep learning models, and achieved impressive performance in Computer Vision~\cite{wang2020contrastive} and Natural Language Processing~\cite{wu2020clear,zhang2021making}.
		Traditionally, CL uses data augmentations to generate ``\textit{positive}'' pairs and randomly selects ``\textit{negative}'' samples from mini-batch without the consideration of labels. 
		The target is to pull together an anchor and a positive sample
		in embedding space, and push apart the anchor from many negative samples~\cite{khosla2020supervised}. 
		This paradigm can be treated as contrastive instance discrimination~\cite{cai2020negatives}. 
		SimCLR~\cite{Chen2020ASF}, MoCo~\cite{he2020momentum}, MAE~\cite{he2021masked} are representative works. 
		
		However, Instance-level CL has some weaknesses in mining common features among instances with the same category. 
		Therefore, researchers have proposed to consider side information for better sampling, such as supervised CL~\cite{khosla2020supervised,gao2021simcse} and Cluster-based CL~\cite{li2021contrastive}. 
		The most relevant work is supervised CL with the consideration of labels. 
		For example, Khosla \textit{et al.}~\cite{khosla2020supervised} used labels to select positive examples with the same labels as the anchor example. Therefore, models can obtain multiple positive examples not only from data augmentation, but also from the same category. 
		One step further, Gao \textit{et al.}~\cite{gao2021simcse} leveraged ``\textit{Entailment}'' and ``\textit{Contradiction}'' labels to select positive examples and negative examples simultaneously. 
		Besides, Yang \textit{et al.}~\cite{yang2021partially,yang2022robust} developed a noise-robust contrastive learning loss for handling the false negative situations in CL for performance improvement.
		In our preliminary work~\cite{zhang2021making}, we designed a relation of relation task to force the model to measure the label relations with contrastive learning.
		They have all made some progress in improving CL performance, and achieved promising results in downstream tasks.
		Nevertheless, these supervised CL-based methods still treat labels as one-hot signals and use these signals to directly select hard examples, underestimating the potential semantic information of labels.
		That is to say, there still is plenty of space for further improving text representation and classification with better label utilization methods.

		\begin{figure*}
			\centering
			\includegraphics[width=0.83\textwidth]{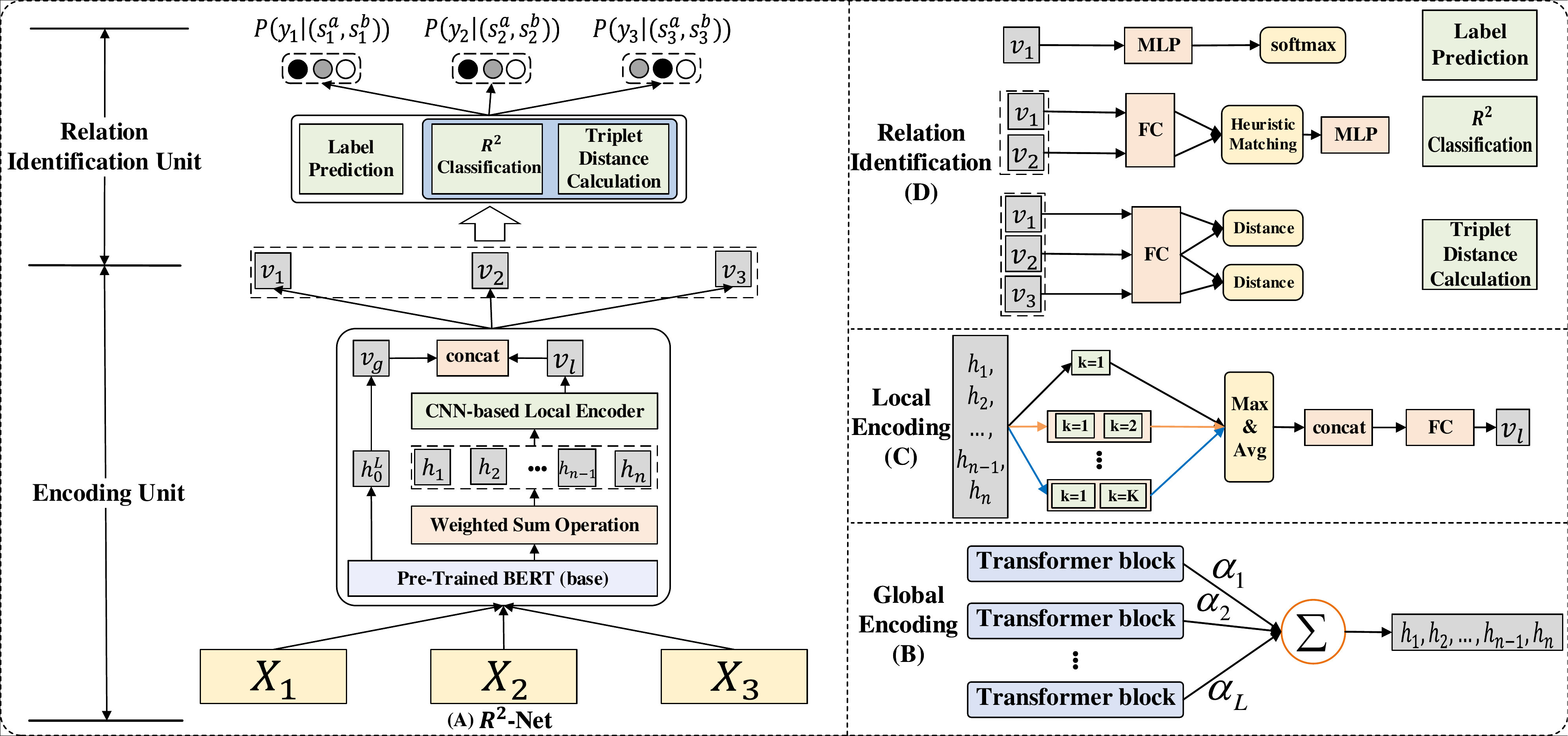}
			\caption{(A) The overall structure of \cname. (B) Using PLMs to obtain the representations of input words and sentences. (C) Our proposed CNN-based local encoding for the model ability enhancement on sentence partial information extraction. (D) Relation identification module for our proposed R$^2$ classification. } 
			\label{f:model}
		\end{figure*}
		
		\section{Problem Statements}
		\label{s:problem}
		
		In this section, we will introduce the definition of text classification task, our proposed relation of relation~(R$^2$) classification task, and necessary notations. 
		
		\subsection{Text Classification}
		Given a word sequence denoted by $\bm{X} = \{x_1, x_2,$ $..., x_n\}$, where $n$ is the length of the word sequence, as well as the label set $\mathcal{Y} (|\mathcal{Y}|=m)$, where the $j^{th}$ label is represented with one-hot representation $\bm{y}_j$. $m$ is the number of labels.
		The target is to learn a classifier $\xi$, which is capable of computing the conditional probability $P(\bm{y}|\bm{X}, \mathcal{Y})$ and predicting the most appropriate label for the input text: 
		\begin{equation}
			\label{eq:task}
			\begin{split}
				&P(\bm{y}|\bm{X}, \mathcal{Y})= \xi(\bm{X}, \mathcal{Y}), \\
				&y^{*} =argmax_{y\in\mathcal{Y}}P(\bm{y}|\bm{X}, \mathcal{Y}),
			\end{split}
		\end{equation}
		where true label $y\in\mathcal{Y}$ indicates the semantic category of input text. 
		For example, $\mathcal{Y}=\{Yes, No\}$ for PI task, and $\mathcal{Y}=\{entailment, contradiction, neutral\}$ for NLI task.

		\subsection{Relation of Relation (R$^2$) Classification}
		Previous study~\cite{gururangan2018annotation} has demonstrated that categories can be helpful to reveal some implicit patterns for text classification. 
		Therefore, we propose a novel \textbf{R}elation of \textbf{R}elation~(R$^2$) classification task to guide models to exploit category information precisely. 
		Given two input texts $\bm{X}_1$ and $\bm{X}_2$, the goal is to learn a classifying function $\mathcal{F}$ to identify whether these two input texts have the same semantic relation:
		\begin{equation}
			\label{eq:r2-task}
			\begin{split}
				\mathcal{F}(\bm{X}_1, \bm{X}_2) = 
				\begin{cases} 
					1,  & \mbox{if } \bm{y}_1 = \bm{y}_2, \\
					0, & \mbox{if } \bm{y}_1 \not= \bm{y}_2,
				\end{cases}
			\end{split}
		\end{equation} 
		where $\bm{y}_1$ and $\bm{y}_2$ stand for one-hot representations of labels of two input texts, respectively. 
		
		Next, we will introduce the technical details of our proposed \cname~and \jname. Some necessary notations are listed in Table~\ref{t:notation} for better illustration.
		
		\begin{table}
			\centering
			\caption{Notations used in \cname~and \jname.}
			\begin{footnotesize}
				\begin{tabular}{c|l} \hline
					\textbf{Notation} & \textbf{Explanation} \\ \hline
					$\bm{X}$ & The input text sequence \\ 
					$\mathcal{Y}$ & The label set \\
					$\bm{H}$ & Word embedding matrix for entire input text \\
					$\bm{v}_g, \bm{v}_l$ & Global and local representations for input text \\ 
					$\bm{v}_*$ & Different vector representations from different modules \\ 
					$\alpha_1, \alpha_2,...,\alpha_L$ & Trainable weights for different layers in PLMs  \\
					$S^{(i)}$ & Fine-grained description set for the $i^{th}$ label \\
					$\bm{E}_f$ & Vanilla label embedding matrix  \\
					$\bm{E}$ & Final label embedding matrix \\
					$\bm{W}_*, \bm{U}_*, \bm{\omega}_*, \bm{b}_*$ & Trainable parameters of our proposed models \\ 
					$\bm{\beta}^l, \bm{\beta}^t$ & Attention weights used in \jname \\
					\hline
				\end{tabular}
			\end{footnotesize}
			\label{t:notation}
		\end{table}

		\section{\fcname~(\cname)}
		\label{s:cmodel}
		
		Fig.~\ref{f:model}(A) reports the overall architecture of \cname.
		To better describe the technical details of \cname, similar to Section~\ref{s:problem}, we elaborate them from two aspects: 1) Text Classification Part; 2) R$^2$ Classification Part. 
		
		\subsection{Text Classification Part}
		\label{s:encode_unit}
		This part focuses on identifying the most suitable label for a given text. 
		Specifically, we first utilize powerful PLMs~\cite{devlin2018bert,Liu2019RoBERTaAR}, such as BERT, to generate sentence semantic representation globally. 
		Meanwhile, we develop a CNN-based encoder to capture keywords and phrase information from a local perspective. 
		Thus, the input text sequence can be encoded in a comprehensive manner. 
		Then, we leverage a Multi-Layer Perceptron~(MLP) to predict the corresponding label. 
		
		\subsubsection{\textbf{Global Encoding}}
		With the full usage of large corpus and multi-layer transformers, PLMs (e.g., BERT~\cite{devlin2018bert}) have accomplished much progress in many NLP tasks. 
		Thus, we take pre-trained BERT as an example to describe how to generate sentence semantic representations for the input. 
		Moreover, inspired by ELMo~\cite{Peters2018DeepCW}, we also use the weighted sum of all the hidden states of words from different transformer layers as the final contextual representations of input words in sentences. 
		
		Specifically, we first split input text into BPE tokens~\cite{sennrich2015neural} and add special token ``\textit{[CLS]}'' at the beginning and the end of word sequence.
		``\textit{[SEP]}'' token is also considered when there is more than one sentence.
		Then, BERT is utilized to generate word representation $\bm{H}$  and sentence representations $\bm{v}_g$.
		As illustrated in Fig.~\ref{f:model}(B), suppose there are $L$ layers in BERT. 
		Contextual word representations for input text sequence $\bm{X}$ is then a pre-layer weighted sum of transformer block output, with weights $\alpha_1, \alpha_2,...,\alpha_L$.
		\begin{equation}
			\label{eq:global-encoding}
			\begin{split}
				&\bm{h}_0^l, \bm{H}^l = BERT_l(\bm{X}), \\
				&\bm{H} = \sum_{l=1}^L\alpha_l\bm{H}^l, \quad \bm{v}_g = \bm{h}_0^L, \\
			\end{split}
		\end{equation} 
		where $\bm{h}_0^l$ denotes the representation of first token ``[\textit{CLS}]'' at the $l^{th}$ layer, and $\bm{v}_g$ denotes the global semantic representation of the input. 
		$\bm{H}$ represents the sequence features of the whole input. $\alpha_l$ is the weight of the $l^{th}$ layer in BERT and will be learned during model training.

		\subsubsection{\textbf{Local Encoding}}
		The semantic relation within the text sequence is not only connected with the important words, but also affected by the local information~(e.g., phrase and local structure). 
		Though BERT leverages multi-layer transformers to perceive important words in the sentence pair, it still has some weaknesses in modelling local information. 
		To alleviate these shortcomings, we develop a CNN-based local encoder to extract local information from the input.  
		
		Fig.~\ref{f:model}(C) reports the local encoding structure. 
		The input of this module is $\bm{H}$ from global encoding. 
		We use convolution operations with different composite kernels~(e.g., bi-gram and tri-gram) to process these features. 
		Each operation with different kernels can model local patterns with different sizes. 
		Next, we leverage average pooling and max pooling to enhance these local features and concatenate them before sending them to a non-linear transformation. 
		Suppose we have $K$ different kernel sizes. This process can be formulated as follows:
		\begin{equation}
			\label{eq:local-encoding}
			\begin{split}
				\bm{H}^k &= CNN_k(\bm{H}), \quad k=1,2,...,K,\\
				\bm{h}^k_{max} &= max(\bm{H}^k), \bm{h}^k_{avg} = avg(\bm{H}^k), \\
				\bm{h}_{concat} &= [\bm{h}^1_{max};\bm{h}^1_{avg};...;\bm{h}^K_{max};\bm{h}^K_{avg}], \\
				\bm{v}_l &= ReLu(\bm{W}\bm{h}_{concat} + \bm{b}), \\
			\end{split}
		\end{equation}
		where $CNN_k$ denotes the convolution operation with the $k^{th}$ kernel. $[\cdot;\cdot]$ is the concatenation operation. $\bm{v}_l$ represents local semantic representation of the input. $\{\bm{W} \in \mathcal{R}^{d_{p}*(2K*d_{p})}, \bm{b}\in \mathcal{R}^{d_{p}}\}$ are trainable parameters. 
		$d_{p}$ is output size of pre-trained model.
		$ReLu(\cdot)$ is the activation function. 
		
		After getting the global representation $\bm{v}_g$ and local representation $\bm{v}_l$, we investigate different fusion methods to integrate them, including simple concatenation, weighted concatenation, as well as weighted sum. Finally, we obtain that simple concatenation is flexible and can achieve comparable performance without extra training parameters. 
		Thus, we employ concatenation $\bm{v} = [\bm{v}_g;\bm{v}_l]$ as the final semantic representation of input text sequence.

		\subsubsection{\textbf{Label Prediction}}
		This component is adopted to predict the label of input text, which is an essential text classification part. 
		To be specific, the input of this component is semantic representation $\bm{v}$. We leverage a two-layer MLP to make the final classification, which can be formulated as follows:
		\begin{equation}
			\label{eq:label-prediction}
			\begin{split}
				P(y|\bm{X}) = MLP_1(\bm{v}). \\
			\end{split}
		\end{equation}

		\subsection{Relation of Relation Learning Part}
		\label{s:predict_unit}
		This part aims to properly and fully use comparison and contrastive learning of label information, and help to improve the model performance. 
		To achieve this goal, we employ two critical modules to analyze \textit{pairwise relation} and \textit{triplet-based relation} simultaneously.

		\subsubsection{\textbf{R$^2$ Classification}}
		\label{s:r2_unit}
		Inspired by SSL methods in PLMs (e.g., MLM and NSP in BERT), we intend \cname~to make full use of label information in a similar way. 
		Therefore, we develop a novel self-supervised R$^2$ classification task. 
		Instead of just identifying the most suitable label, 
		we plan to obtain more information about the input text sequence by analyzing the \textit{pairwise relation} between semantic representations (i.e., $\bm{v}_1$ for $\bm{X}_1$, and $\bm{v}_2$ for $\bm{X}_2$). 
		Since a learnable nonlinear transformation between representations and loss substantially improves the model performance~\cite{Chen2020ASF}, we first transfer $\bm{v}_1$ and $\bm{v}_2$ with a nonlinear transformation. 
		Then, we leverage heuristic matching~\cite{Chen-Qian2017ACL,zhang2017context} to model the similarity and difference between $\bm{v}_1$ and $\bm{v}_2$. 
		Next, we send the matching result $\bm{u}$ to an MLP with one hidden layer for final classification. This process is formulated as follows:
		\begin{equation}
			\label{eq:r2-prediction}
			\begin{split}
				&\bar{\bm{v}}_1 = MLP_1(\bm{v}_1), \quad
				\bar{\bm{v}}_2 = MLP_1(\bm{v}_2), \\
				&\bm{u} = [\bar{\bm{v}}_1; \bar{\bm{v}}_2; (\bar{\bm{v}}_1 \odot \bar{\bm{v}}_2); (\bar{\bm{v}}_1 - \bar{\bm{v}}_2)], \\
				&P(\hat{y}|\bm{X}_1, \bm{X}_2) = MLP_2(\bm{u}), \\
			\end{split}
		\end{equation}
		where concatenation can retain all the information~\cite{zhang2017context}. The element-wise product is a certain measure of ``similarity'' between two sentences~\cite{mou2016natural}. Their differences can capture the degree of distributional inclusion in each dimension~\cite{weeds2014learning}. 
		$\hat{y}\in\{1, 0\}$ indicates whether two input sequences have the same relation.

		\subsubsection{\textbf{Triplet Distance Calculation}}
		Apart from leveraging R$^2$ task to learn inter-class relation information, we also intend to learn intra-class connections from \textit{triplet-based relation}. 
		Thus, we introduce triplet loss~\cite{Schroff2015FaceNetAU} into \cname. 
		As a fundamental similarity function, triplet loss is widely applied in information retrieval~\cite{Liu2010LearningTR}, and is able to pull together input sequences with the same label and push apart these with different labels. 
		The inputs of this component are three semantic representations: 
		$\bm{v}_a$ for anchor text $\bm{X}_a$, $\bm{v}_p$ for positive text $\bm{X}_p$, $\bm{v}_n$ for negative text $\bm{X}_n$. 
		We first transform them into a common space with a full connection layer~\cite{Chen2020ASF}. 
		Then, we calculate the distance between anchor and positive, as well as the distance between anchor and negative, respectively: 
		\begin{equation}
			\label{eq:distance-prediction}
			\begin{split}
				& \bar{\bm{v}}_i = ReLu(\bm{W}_d\bm{v}_i + \bm{b}_d), \quad i\in\{a, p, n\}, \\
				& d_{ap} = Dist(\bar{\bm{v}}_a, \bar{\bm{v}}_p), \quad
				d_{an} = Dist(\bar{\bm{v}}_a, \bar{\bm{v}}_n), \\
			\end{split}
		\end{equation}
		where $\{\bm{W}_d\in \mathcal{R}^{d_m*d_p},  \bm{b}_d\in \mathcal{R}^{d_m}\}$ are trainable parameters. 
		$d_m$ is the hidden state size. 
		$Dist(\cdot)$ is the distance calculation function, which we use \textit{Euclidean} distance.

		\begin{figure}
			\centering
			\includegraphics[width=0.4\textwidth]{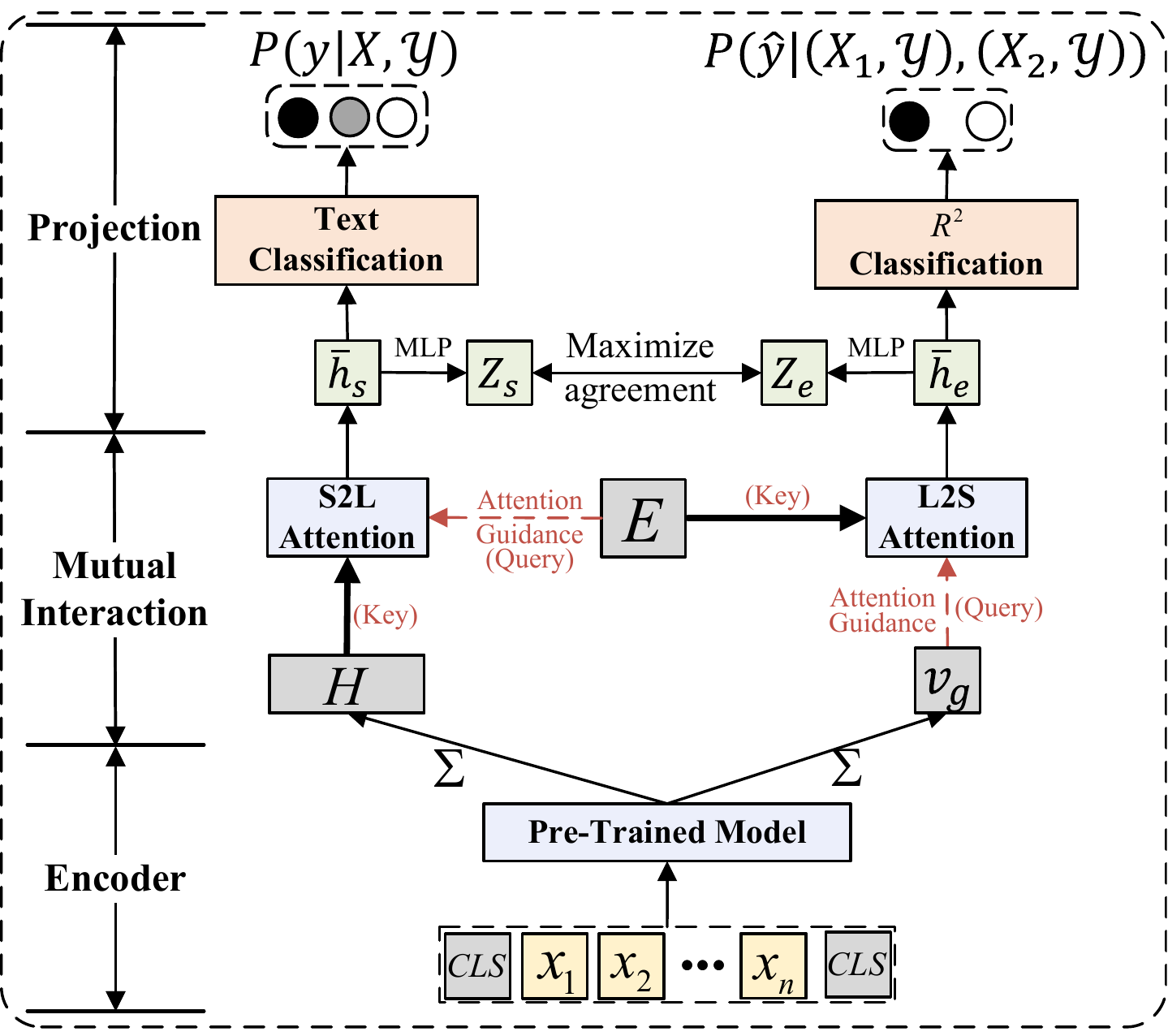}
			\caption{The overall structure of \jname, which consists of \textit{Encoder}, our proposed \textit{Mutual Interaction}, and \textit{Projection}. (Query) and (Key) denote the components in attention calculation.}
			\label{f:dele}
		\end{figure}
		
		\subsection{Model Learning}
		As mentioned in Section~\ref{s:problem}, both text classification and R$^2$ task can be treated as classification tasks. 
		Thus, we employ \textit{Cross-Entropy} as the loss for each input as follows:
		
		\begin{equation}
			\label{eq:loss_classify}
			\begin{split}
				L_{s} &= -\sum_{i=1}^{N}\bm{y}_i \mathrm{log} P(y_i | \bm{X}_i), \\ 
				L_{R^2} & = -\sum_{j=1}^{N/2}\hat{\bm{y}}_{j} \mathrm{log} P(\hat{y}_j|(\bm{X}_1, \bm{X}_2)_j), \\
			\end{split}
		\end{equation}
		where $N$ is the number of one training batch. $\bm{y}_i$ is the one-hot vector for the true label of the $i^{th}$ instance. $\hat{\bm{y}}_{j}$ is the one-hot vector for the true relation of relations of the $j^{th}$ instance pair. 
		
		Moreover, we introduce triplet loss to better analyze the connections and differences among labels of different pairs:
		\begin{equation}
			\label{eq:loss_dist}
			\begin{split}
				L_{d} = \sum_{i=1}^{N/3}max((d_{ap}-d_{an}+margin)_i, 0). \\
			\end{split}
		\end{equation}
		
		Finally, we treat the weighted sum of these losses with a hyper-parameter $\eta$ as the loss function for one mini-batch:
		\begin{equation}
			\label{eq:loss}
			\begin{split}
				Loss = \eta L_{s} + (1-\eta)(L_{R^2} + L_{d}). \\
			\end{split}
		\end{equation}

		\section{\fjname~(\jname)}
		\label{s:jmodel}
		In our preliminary work~\cite{zhang2021making}, we propose \cname~to exploit label information in a one-hot manner for better supervision and text classification. 
		However, one-hot usage still needs to improve in exploiting labels comprehensively. 
		As mentioned in Section~\ref{s:introduction}, representing labels with dense vectors can be an inspiring direction. 
		However, the particular meaning required in label semantics and multiple semantics contained in label words or phrases pose a big challenge for label embedding methods.
		Thus, in this section, we focus on describing and representing label semantics in a fine-grained manner.
		
		Specifically, we propose to employ fine-grained descriptions to generate better label embedding, and extend \cname~to a novel \jname. 
		The overall architecture is reported in Fig.~\ref{f:dele}.
		Key contributions lie in the following two areas. 
		The first is incorporating fine-grained descriptions from WordNet to enrich the label embedding learning process, so that label semantics can be fully exploited. 
		The second is developing a novel mutual interaction module based on CL framework, which is used to analyze bidirectional interactions between input text sequences and labels.  
		Along this line, label semantics can better guide the importance selection from input text. 
		Meanwhile, text representations can be used to denoise fine-grained descriptions for better label embedding learning. 
		
		As reported in Fig.\ref{f:dele}, \jname~consists of three modules: 1) \textit{Encoder} module: encoding input text sequence and all labels; 2) \textit{Mutual Interaction} module: developing S2L Attention and L2S Attention to enhance embedding learning and denoise fine-grained descriptions; 3) \textit{Projection} module: projecting different learned embeddings into same space for contrastive learning. 
		Next, we will report the technical details of \jname. 
		
		\subsection{Encoder}
		For input text sequence, we employ a similar operation to \cname~to encode the input text sequence, and obtain the word representations $\bm{H}$ and global sequence representation $\bm{v}_g$.

		As for labels, we design a novel \textit{Label Encoder} to incorporate fine-grained descriptions for label embedding learning. 
		Specifically, we employ WordNet as prior knowledge and retrieve relevant descriptions for labels.
		Since noun senses of words are usually employed as labels, we select noun descriptions of labels from WordNet.
		Moreover, words in WordNet have about 25 attributes (e.g., noun.act, noun.animal, noun.event)\footnote{https://wordnet.princeton.edu/documentation/lexnames5wn}. 
		To ensure the relevance and mitigate the noise problem, we manually select no more than 3 noun descriptions for each label word based on most relevant attributes.
		If the label consists of more than one word (e.g., ``\textit{Business \& Finance}'' in Yahoo! Answer dataset), we select no more than 3 noun descriptions for each word and treat all of them as fine-grained descriptions for the label.
		After that, our designed label encoder is used to generate the corresponding representation.  
		
\begin{figure}
			\centering
			\includegraphics[width=0.4\textwidth]{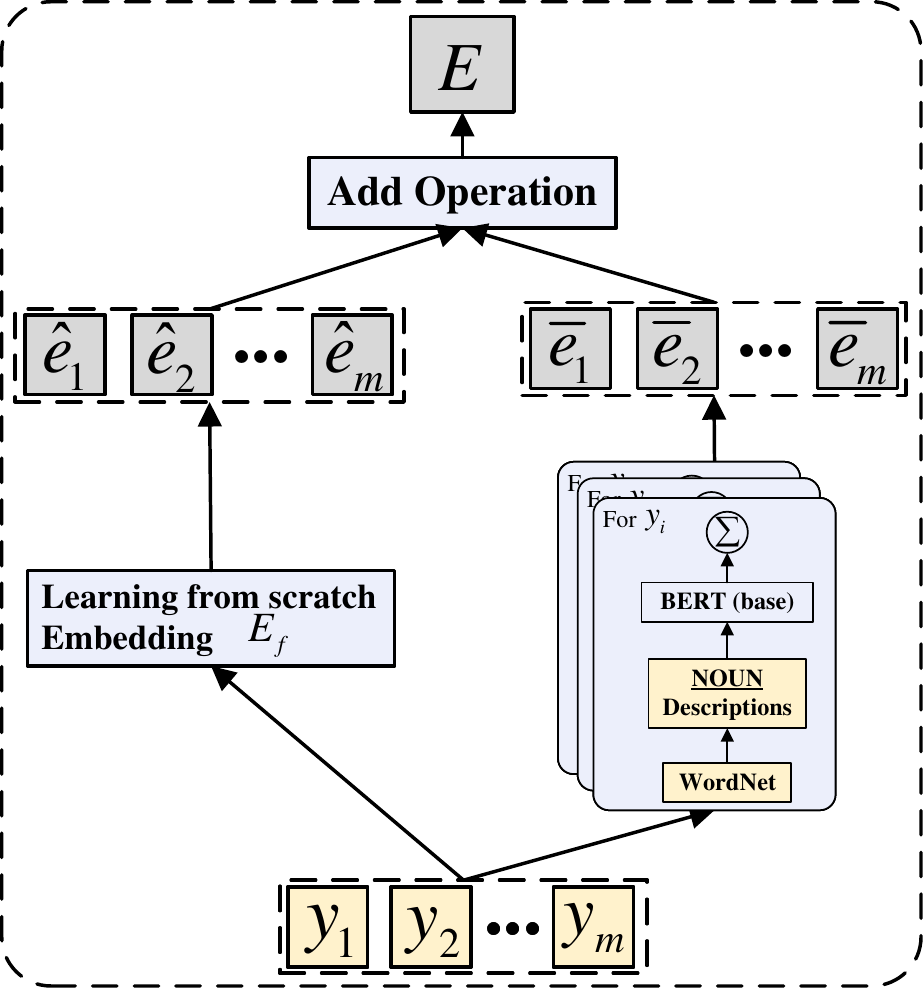}
			\caption{The Architecture of Label Encoder in \jname, which incorporates vanilla embedding and our proposed description-enhanced embedding.}
			\label{f:label_encoder}
		\end{figure}

		Fig.~\ref{f:label_encoder} demonstrates the structure of label encoder.
		Taking the $i^{th}$ label as an example, we first use $\bm{E}_f\in\mathbb{R}^{m*d_p}$ to represent the vanilla label embedding matrix.
		The embedding of the $i^{th}$ label can be obtained by extracting the $i^{th}$ column $\hat{\bm{e}}_i$ from $\bm{E}_f$.
		Meanwhile, we obtain fine-grained description set $\bm{S}^{(i)}$ for the $i^{th}$ label, where the $j^{th}$ description can be denoted as $\bm{s}^{(i)}_j$.
		Then, we employ BERT to encode these descriptions and average the output of ``\textit{[CLS]}'' from the last layer as the description representation.
		Next, we add the vanilla embedding $\hat{\bm{e}}_i$ and the description representation $\bar{\bm{e}}_i$ to get the enriched representation $\bm{e}_i$ for the $i^{th}$ label as follows:
		\begin{equation}
			\label{eq:label_encoder}
			\begin{split}
				\bar{\bm{e}}_i &= \frac{1}{|\bm{S}^{(i)}|}\sum_{j=1}^{|\bm{S^{(i)}}|}BERT_L(\bm{s}^{(i)}_j), \\
				\hat{\bm{e}}_i &= \bm{y}_i \bm{E}_f, \quad \bm{e}_i = \hat{\bm{e}}_i + \bar{\bm{e}}_i. \\
			\end{split}
		\end{equation}
		
		\subsection{Mutual Interaction}
		In the previous part, we introduced fine-grained descriptions from WordNet to enrich the label embedding learning.
		However, only the specific meaning of a word is treated as a label.
		Thus, not all descriptions are helpful, and some even obscure label semantics.
		To this end, we develop a novel mutual interaction module based on CL to mitigate this unexpected noise.
		As shown in Fig.~\ref{f:dele}, this module leverages \textit{S2L Attention} and \textit{L2S Attention} to tackle this problem.
		
		\subsubsection{S2L Attention} 
		Since label semantic meaning is highly related to input text, we first leverage text representation $\bm{v}_g$ as the guidance to select necessary label representations from label embedding $\bm{E}$ as follows:
		\begin{equation}
			\label{eq:label_semantics}
			\begin{split}
				\bm{E} &= [e_1, e_2, ..., e_m], \\
				\bm{\beta}^l &= \bm{\omega}_l^Ttanh(\bm{W}_l\bm{E}+\bm{U}_l\bm{v}_g\otimes\bm{I}_l), \\
				\bar{\bm{h}}_e &= \sum_{i=1}^{m}\frac{exp(\beta^l_i)}{\sum_{k=1}^{m}exp(\beta^l_k)}e_i,
			\end{split}
		\end{equation}
		where $\{\bm{\omega}_l \in \mathcal{R}^{1*d_a}, \bm{W}_l \in \mathcal{R}^{d_a*d_p}, \bm{U}_l\in \mathcal{R}^{d_a*d_p}\}$ are trainable parameters.  $d_a$ is the size of attention unit.
		$\bm{I}_l\in\mathbb{R}^m$ is a row vector of $1$.
		$\bm{U}_l\bm{v}_g\otimes\bm{I}_l$ means repeating $\bm{U}_l\bm{v}_g$ m times.
		$\bm{\beta}^l$ is the unnormalized attention weight for label representations.
		$\bar{\bm{h}}_e$ denotes the text supervised semantic vector generated from label semantics.
		With this operation, the unexpected noise from fine-grained descriptions can be effectively mitigated and label semantics can be expressed accurately.
		
		\subsubsection{L2S Attention} 
		Meanwhile, we intend to analyze the impact from label to text in a similar way.
		Since there are multiple labels, we adopt each label to guide the selection of input text.
		Then, max-pooling is employed to merge results for label supervised semantic vectors.
		\begin{equation}
			\label{eq:text_semantics}
			\begin{split}
				\beta^t &= \bm{\omega}^Ttanh(\bm{W}\bm{H}+\bm{U}\bm{e}_t\otimes\bm{I}), \\
				\bar{\bm{h}}_s^t &= \sum_{i=1}^{n}\frac{exp(\beta^t_i)}{\sum_{k=1}^{n}exp(\beta^t_k)}\bm{h}_i, \quad \bm{h}_i \in \bm{H}  \\
				\bar{\bm{h}}_s &= maxpooling([\bar{\bm{h}}_s^1, \bar{\bm{h}}_s^2, ..., \bar{\bm{h}}_s^m,]) ,
			\end{split}
		\end{equation}
		where $t\in\{1, 2, ..., m\}$ indicates the index of the $t^{th}$ label.
		$\beta^t$ is the unnormalized attention weight for word representations under the guidance of the $t^{th}$ label.
		$\{\bm{\omega}\in \mathcal{R}^{1*L_s}, \bm{W}\in \mathcal{R}^{L_s*d_p}, \bm{U}\in \mathcal{R}^{L_s*d_p}\}$ are also trainable parameters. $L_s$ is the sentence length. $\bar{\bm{h}}_s$ denotes the label supervised semantic vector from sentence semantics.

		\subsection{Projection.}
		In the aforementioned sections, we have obtained semantic vectors from the text perspective and label perspective.
		To minimize their distance for better representation learning and classification, inspired by SimCLR~\cite{chen2020simple}, we employ an MLP to project the semantic vectors into the same space for quality improvement of learned semantic vectors as follows:
		\begin{equation}
			\label{eq:projection}
			\begin{split}
				\bm{z}_e = MLP_2(\bar{\bm{h}}_e), \quad
				\bm{z}_s = MLP_2(\bar{\bm{h}}_s), \\
			\end{split}
		\end{equation}
		where $\bm{z}_e$ and $\bm{z}_s$ are projection vectors for the contrastive loss.
		Since our target is to predict the most suitable label for input text and $\bm{h}_s$ is generated from the text perspective,  we select label supervised semantic vector $\bar{\bm{h}}_s$ to make the final decision:
		\begin{equation}
			\label{eq:prediction}
			\begin{split}
				&P(y|\bm{X}, \mathcal{Y}) = MLP_3(\bar{\bm{h}}_s), \\
				&y^{*} =argmax_{y\in\mathcal{Y}}P(\bm{y}|\bm{X}, \mathcal{Y}).
			\end{split}
		\end{equation}
		
		Similarly, we also employ R$^2$ classification task to assist in the improvement of model ability. Text supervised semantic vector $\bm{h}_e$ is selected to make the prediction as follows: 
		\begin{equation}
			\label{eq:r2_prediction}
			\begin{split}
				&\hat{\bm{u}} = [\bar{\bm{h}}_e^1; \bar{\bm{h}}_e^2; (\bar{\bm{h}}_e^1\odot\bar{\bm{h}}_e^2);(\bar{\bm{h}}_e^1-\bar{\bm{h}}_e^2)], \\
				&P(\hat{y}|(\bm{X}_1,\mathcal{Y}), (\bm{X}_2, \mathcal{Y})) = MLP_4(\hat{\bm{u}}), \\
				&\hat{y}^{*} =argmax_{\hat{y}\in\{0,1\}}P(\hat{y}|(\bm{X}_1,\mathcal{Y}), (\bm{X}_2, \mathcal{Y})),
			\end{split}
		\end{equation}
		where $MLP_3(\cdot)$ and $MLP_4(\cdot)$ are MLPs, which consist of one hidden layer and a softmax output layer. $y^{*}$ and $\hat{y}^{*}$ are the predicted label for $\bm{X}$ and predicted relation of relations of $\bm{X}_1$ and $\bm{X}_2$. 
		
		\begin{table*}
			\centering
			\caption{Experimental Results~(accuracy) on different datasets from NLI task. \\ $+$ and $-$ denote the percentage of decrease or increase in \textbf{error rate} compared with backbones.}
			\begin{tabular}{c|lcccc} \hline
				\textbf{Type} & \textbf{Model} & \textbf{\tabincell{c}{SNLI Full Test\\(3-classes)}} & \textbf{\tabincell{c}{SNLI Hard Test\\(3-classes)}} & \textbf{\tabincell{c}{SICK Test \\(3-classes)}} & \textbf{\tabincell{c}{SciTail Test\\(2-classes)}} \\ \hline
				\multirow{9}{*}{\tabincell{c}{One-hot\\Encoding}}
				&(1) DRCN~\cite{Kim2018SemanticSM} & 86.5\%& 68.3\% & 87.4\% & 85.7\%\\
				&(2) CSRAN~\cite{tay2018co} & 88.5\% & 76.8\% & 89.7\% & 86.5\%\\
				&(3) RE2~\cite{yang2019simple} & 88.9\% & 77.3\% & 89.8\% & 86.2\%\\
				&(4) DRr-Net~\cite{zhang2019drr} & 87.7\% & 71.4\% & 88.3\% & 87.4\%\\ 
				&(5) SimCSE~\cite{gao2021simcse} & 89.4\% & 81.2\% & 90.3\% & 93.3\%\\
				\cline{2-6}
				&(6) BERT-(base)~\cite{devlin2018bert} & 90.3\%~(0.0\%) & 80.6\%~(0.0\%) & 88.7\%~(0.0\%) & 93.1\%~(0.0\%) \\
				&(7) BERT-(large)~\cite{devlin2018bert} & 90.7\%~($+$4.12\%) & 81.3\%~($+$3.61\%) & 88.3\%~($-$3.54\%) & 93.6\%~($+$7.25\%) \\
				&(8) RoBERTa-(base)~\cite{Liu2019RoBERTaAR} & 90.9\%~(0.0\%) & 81.5\%~(0.0\%) & 90.3\%~(0.0\%) & 93.8\%~(0.0\%) \\
				&(9) ALBERT-(base)~\cite{Lan2020ALBERTAL} & 86.2\% & 77.5\%& 87.3\% & 91.4\%\\ \hline
				\multirow{3}{*}{\tabincell{c}{Label\\Embedding}}
				& (10)FLE-BERT(base)~\cite{xiong2021fusing} &90.5\%~($+$2.06\%) & 80.4\%~($-$1.03\%) & 89.3\%~($+$5.31\%) & 93.4\%~($+$4.35\%) \\
				&(11) LEAM-BERT(base)~\cite{wang2018jointembedding} & 90.3\%~(0.0\%) & 80.8\%~($+$1.03\%) & 89.0\%~($+$2.65\%) & 93.4\%~($+$4.35\%) \\
				&(13) EXAM-BERT(base)~\cite{du2019explicit} &  90.6\%~($+$3.09\%) & 81.0\%~($+$2.06\%) & 89.5\%~($+$7.08\%) & 93.8\%~($+$10.14\%) \\
				& (12) LGDSC-BERT(base)~\cite{zhu2022generating} & 91.3\%~($+$10.31\%) & 81.4\%~($+$4.12\%) & 89.5\%~($+$7.08\%) & 94.0\%~($+$13.04\%) \\
				& (14) LCM-BERT(base)~\cite{guo2020label} & 90.8\%~($+$5.15\%) & 81.3\%~($+$3.61\%) & 90.3\%~($+$14.16\%) & \underline{94.1}\%~($+$14.49\%) \\
				&(15) EXAM-RoBERTa(base)~\cite{du2019explicit} &  \underline{91.5\%}~($+$6.59\%) & \underline{81.9\%}~($+$2.16\%) & \underline{90.5\%}~($+$2.06\%) & \underline{94.1\%}~($+$4.84\%) \\ 
				\hline
				\multirow{4}{*}{\tabincell{c}{Our\\Methods}}
				&(16) \cname-BERT-(base) & 91.1\%~($+$8.25\%) & 81.0\%~($+$2.06\%) & 89.2\%~($+$4.42\%) & 92.9\%~($-$2.89\%) \\
				&(17) \cname-RoBERTa-(base) & 91.3\%~($+$4.40\%) & 81.4\%~($-$0.54\%) & 89.5\%~($-$8.24) & 93.9\%~($+$1.61) \\
				&(18) \jname-BERT-(base) & 91.3\%~($+$10.31\%) & 81.6\%~($+$5.15\%) & 89.9\%~($+$10.16\%) & \underline{94.1\%}~($+$14.49\%)\\
				&(19) \jname-RoBERTa-(base) & \textbf{91.8\%}~($+$9.89\%) & \textbf{83.2\%}~($+$9.19\%) & \textbf{90.7\%}~($+$4.12\%) & \textbf{94.5\%}~($+$11.29\%) \\
				\hline
			\end{tabular}
			\label{t:nli-result}
		\end{table*}
		
		\subsection{Model Learning}
		\jname~has three targets to optimize, including \textit{contrastive learning target}, \textit{R$^2$ classification target}, and \textit{text classification target}. They have been listed in the following part.

		\textbf{Contrastive Learning Target.}
		As mentioned before, minimizing the distance between semantic representations of input text and the proper label is pivotal for text classification.
		Thus, we select \textit{NT-Xent}~\cite{chen2020simple} as the contrastive loss, where the positive pair consists of \textit{label supervised semantic vector} and \textit{text supervised semantic vector}, and negative examples are the rest instances in current batch:
		\begin{equation}
			\begin{split}
				L_1 = \sum_{i=1}^{N} -log\frac{sim(\bm{z}^i_s, \bm{z}^i_e)/\tau}{\sum_{j=1}^{K}\mathbbm{1}_{[j\ne i]}sim(\bm{z}^i_s, \bm{z}^j_s)/\tau}, \\
			\end{split}
		\end{equation}
		where $sim(\cdot)$ is the similarity calculation function. $N$ is the number of one training batch. $\tau$ is the temperature parameter. $\mathbbm{1}_{[j\ne i]}$ is an indicator function evaluating to 1 iff $j\ne i$.
		
		\textbf{R$^2$ classification target.}
		For this classification target, we select \textit{Cross-Entropy} as the training loss:
		\begin{equation}
			\begin{split}
				L_2 = \sum_{i=1}^{N/2} -\hat{\bm{y}}_ilogP(\hat{y}_i|(\bm{X}_1,\mathcal{Y})_i, (\bm{X}_2, \mathcal{Y})_i), \\ 
			\end{split}
		\end{equation}
		where  $\hat{\bm{y}}_i$ is the one-hot ground truth that indicates the true relation of relations of the $i^{th}$ pair.
		
		\textbf{Text Classification Target.}
		For this task-related target, we select \textit{Cross-Entropy} as the training loss:
		\begin{equation}
			\begin{split}
				L_3 = \sum_{i=1}^{N} -\bm{y}_ilogP(y_i|\bm{X}, \mathcal{Y}), \\
			\end{split}
		\end{equation}
		where $\bm{y}_i$ is the one-hot ground truth that represents the true label of the $i^{th}$ example.
		
		The final loss for one mini-batch can be calculated with a weighted sum operation.
		Here, we leverage weight $\delta$ and $\mu$ to control the impacts of CL target and R$^2$ task for the final classification performance:
		\begin{equation}
			\label{eq:jloss}
			Loss = \delta L_1 + \mu L_2 + L_3.
		\end{equation}

		\section{Experiments}
		\label{s:experiment}
		
		In this section, the evaluation datasets and metrics are first introduced. Then, model implementation and training details are reported for better illustration. 
		Next, empirical results, a detailed analysis of models, and experimental results are presented. 
		For all reported results, we employ \textbf{boldface} and \underline{underline} for the best and the second-best results, respectively.
		
		\begin{table}
			\centering
			\caption{Hyper-parameters configuration in \cname~and \jname.}
			\begin{footnotesize}
				\begin{tabular}{l|lr} \hline
					\textbf{Model} & \textbf{Hyper-parameters} & \textbf{Value} \\ \hline
					BERT-base & \multicolumn{2}{l}{12 layers, hidden size $d_p=768$, attention heads 12} \\
					Roberta-base & \multicolumn{2}{l}{12 layers, hidden size $d_p=768$, attention heads 12}  \\
					\hline
					\multirow{5}{*}{\cname} 
					& Kernel size of \textit{CNN} & $d_k = \{1,2,3\}$ \\
					& hidden size of $MLP_1$ & $d_m = 300$ \\
					& hidden size of $MLP_2$ & $d_2 = 300$ \\
					& Backbone learning rate & $lr_1 = 10^{-5}$ \\
					& The other learning rate & $lr_2 = 10^{-3}$ \\
					& margin in loss $L_d$ & $margin = 0.2$ \\
					\hline 
					\multirow{4}{*}{\jname} 
					& number of backbone layer & $L = 2$ \\
					& attention size of mutual interaction & $d_a = 100$ \\
					
					& hidden size of $MLP_3$ and $MLP_4$ & $d_3 = 200$ \\
					
					\hline
				\end{tabular}
			\end{footnotesize}
			\label{t:hyper-parameter}
		\end{table}
		
		\begin{table*}
			\centering
			\caption{Results~(accuracy) on PI, Sentiment Analysis, and QA Topic Classification task. \\ $+$ and $-$ denote the percentage of decrease or increase in error rate compared with backbones.}
			\begin{tabular}{c|lcccc} \hline
				\textbf{Type} &\textbf{Model} & \textbf{\tabincell{c}{Quora Test\\(2-classes)}} & \textbf{\tabincell{c}{MSRP Test\\(2-classes)}} &
				\textbf{\tabincell{c}{STS-5 Test\\(5-classes)}} &\textbf{\tabincell{c}{Yahoo! Answer Test\\(10-classes)}}\\ \hline
				\multirow{7}{*}{\tabincell{c}{One-hot\\Encoding}}
				&(1) DRCN~\cite{Kim2018SemanticSM} & 90.2\% & 82.5\% &- &75.1\%\\ 
				&(2) RE2~\cite{yang2019simple} & 89.3\% & 78.5\% & -&75.5\% \\ 
				&(3) DRr-Net~\cite{zhang2019drr} & 89.8\% & 82.9\% & 50.8\% &74.7\% \\ &(4) SimCSE~\cite{gao2021simcse} & 91.5\% & 84.8\% & 54.1\% & 76.2\%\\
				\cline{2-6}
				&(5) BERT-(base)~\cite{devlin2018bert} & 91.0\%~(0.0\%) & 84.2\%~(0.0\%) & 53.1\%~(0.0\%) &75.7\%~(0.0\%)\\ 
				&(6) BERT-(large)~\cite{devlin2018bert} & 91.4\%~($+$4.44\%) & 85.4\%~($+$7.59\%)& 54.9\%~(+3.84\%) &76.3\%~($+$2.67\%)\\ 
				&(7) RoBERTa-(base)~\cite{Liu2019RoBERTaAR} & 90.6\%~(0.0\%) & \underline{87.1\%}~(0.0\%) & 56.2\%~(0.0\%)&75.9\%~(0.0\%)\\ 
				&(8) ALBERT-(base)~\cite{Lan2020ALBERTAL} & 90.3\% & \textbf{88.6\%}& 47.3\% &74.2\% \\ \hline
				\multirow{3}{*}{\tabincell{c}{Label\\Embedding}}
				& (9)FLE-BERT(base)~\cite{xiong2021fusing} & 91.2\%~($+$2.22\%) & 84.2\%~(0.0\%) & 53.4\%~($+$0.64\%) & 75.6\%~($-$0.41\%) \\
				&(10) LEAM-BERT(base)~\cite{wang2018jointembedding} & 91.3\%~($+$3.34\%) & 83.9\%~($-$1.90\%) & 53.8\%~(+1.49\%) &75.9\%~($+$0.82\%)\\
				&(11) EXAM-BERT(base)~\cite{du2019explicit} &  91.4\%~($+$4.44\%) & 84.2\%~(0.0\%) & 54.3\%~(+2.56\%) &76.3\%~($+$2.47\%)\\
				& (12) LGDSC-BERT(base)~\cite{zhu2022generating} & 91.6\%~($+$6.67\%) & 84.5\%~($+$1.90\%) & 54.3\%~($+$2.56\%) & 76.2\%~($+$2.06\%) \\
				& (13) LCM-BERT(base)~\cite{guo2020label} & 91.5\%~($+$5.56\%) & 84.2\%~(0.0\%) & 54.6\%~($+$3.20\%) & 76.4\%~($+$2.88\%) \\
				&(14) EXAM-RoBERTa(base)~\cite{du2019explicit} &  \underline{92.0\%}~($+$14.89\%) & 85.6\%~($-$11.63\%) & \underline{57.0\%}~(+1.82\%) &76.3\%~($+$1.65\%)\\ \hline
				\multirow{4}{*}{\tabincell{c}{Our\\Methods}}
				&(15) \cname-BERT-(base) & 91.6\%~($+$6.67\%) & 84.3\%~($+$0.63\%) & 54.1\%~(+2.13\%)&76.2\%~($+$2.06\%)\\
				&(16) \cname-RoBERTa-(base) & 91.7\%~($+$11.70\%) & 84.5\%~($-$20.15\%) & 56.9\%~(+1.60\%)&76.3\%~($+$1.65\%)\\
				&(17) \jname-BERT-(base) & 91.7\%~($+$7.78\%) & 84.8\%~($+$3.80\%) & 54.7\%~(+3.41\%)&\underline{76.5\%}~($+$3.29\%)\\ 
				&(18) \jname-RoBERTa-(base) & \textbf{92.3\%}~($+$18.09\%) & 86.7\%~($-$3.10\%) &\textbf{57.8\%}~(+3.65\%) &\textbf{76.8\%}~($+$3.73\%)\\
				\hline
			\end{tabular}
			\label{t:pi-qa-result}
		\end{table*}

		\subsection{Datasets and Evaluation Method} 
		To evaluate our proposed \cname~and \jname~comprehensively, we select different benchmark datasets: 
		\textit{SNLI}~\cite{bowman2015large}, \textit{SICK}~\cite{marelli2014semeval}, and \textit{SciTail}~\cite{khot2018scitail} for Natural Language Inference~(NLI), 
		\textit{Quora Question Pair (Quora)}~\cite{iyer2017first} and \textit{MSRP}~\cite{dolan2005automatically} datasets for Paraphrase Identification~(PI), 
		\textit{Yahoo! Answers (Yahoo)} for QA topic classification, 
		as well as \textit{SST-5}~\cite{socher2013recursive} for sentiment classification. 
		These tasks focus on different aspects and exhibit different characteristics of text classification task. 
		
		For evaluation, we select \textit{Accuracy} and \textit{Error Rate Comparison} as evaluation metrics, which are the same as most baselines did. 
		We have to note that for each experiment, we repeat the evaluation process $5$ times with different seeds and random initialization, and report the best results for our proposed methods and baselines. 
		
		\subsection{Model Implementation and Training Details}
		To obtain the best performance, we have tuned hyper-parameters on validation set of each dataset, and used \textit{Early-Stop} operation to select the best values for hyper-parameters. BERT-(base) and Roberta-(base) are selected as backbones. 
		We have to note that all baselines have the same experimental settings for the fair comparison. 
		In order to achieve the best model performance, \cname~and \jname~have different parameter settings over different datasets.
		Therefore, we report common hyper-parameters in Table~\ref{t:hyper-parameter} and summarize them here: 
		
		For \cname, kernel sizes of CNN in local encoding are $d_k=1, 2, 3$. 
		The hidden size of $MLP_1$ is $d_m=300$.
		The margin in Eq.~(\ref{eq:loss_dist}) is $margin=0.2$.
		For PLMs, we set learning rate $10^{-5}$ and use AdamW to fine-tune parameters. 
		For the rest parameters, learning rate is set as $10^{-3}$ and decreases as the model training. 
		An Adam optimizer with $\beta_1 =0.9$ and $\beta_2 = 0.999$ is adopted to optimize these parameters. 
		
		For \jname, the number of used output layers in PLMs is $L=2$. 
		Attention size in mutual interaction module is $d_a=100$. 
		The hidden size of $MLP_2$ in projection is $d_2=300$.
		The hidden size of $MLP_3$ and $MLP_4$ in classification layer is $d_3=200$.
		For model training, we leverage Adam as the optimizer. 
		Inspired by~\cite{howard2018universal}, we develop the following operation to control the learning rate in the $i^{th}$ training batch:
		\begin{equation}
			\small
			\begin{aligned}
				&lr = \epsilon(lr_{max}-lr_{min}) + lr_{min} \\
				&\mu = \left\{
				\begin{aligned}
					&0.5\mathrm{cos}\left(\frac{\pi}{\eta C}(i-\eta C + 1)\right) + 0.5,~if~i \le \eta C, \\
					&\left( 0.5\mathrm{cos}\left(\frac{\pi}{C-\eta C}(i-\eta C)\right) + 0.5 \right)^2,~if~i > \eta C, \\
				\end{aligned}
				\right.
			\end{aligned}
		\end{equation}
		where $lr_{max} = 1$ and $lr_{min} = 0.000001$. $C$ is the total number of training batches, $\eta \in [0, 1]$ is the percentage of the warm up loops.  
		
		\subsection{Model Complexity Analysis}
		In order to better demonstrate the superiority of our proposed methods, we make an additional computational complexity analysis. 
		Since our proposed methods are both based on PLMs, we only analyze the extra computational complexity.
		For our preliminary work \cname, additional components consist of CNN-based local encoder, R$^2$ classification, and triplet distance calculation components. 
		For CNN-based local encoder, we leverage three kernels with size $d_k = \{1, 2, 3\}$ and an MLP. The extra parameter size is $14+d_p*(6*d_p)$. 
		For the rest two components, three different MLPs are used. Therefore, the extra parameter size is $(d_m*2d_p) + (d_p*4d_m) + (d_m*d_p) $. 
		Therefore, total extra computational complexity increase is $(7d_p + 6d_m) * d_p + 14$.
		
		For our proposed \jname, additional components include mutual interaction module and R$^2$ classification layer. 
		For the former, the extra parameter size is $(d_a + d_a * d_p + d_a * d_p) + (L_s + L_s*d_p + L_s * d_p)$.
		For the latter, the extra parameter size is $d_4 * 4d_p + d_4$.
		To this end, total extra computational complexity increase is $(2d_a + 2L_s + 4d_4)*d_p + (d_a + L_s + d_4)$.
		Note that $L_s$ is the sentence length and other notation values can be found in Table~\ref{t:hyper-parameter}. 
		For our proposed label encoder, since we leverage the same PLMs to encoder the fine-grained descriptions as input encoder, this label encoder does not add additional computational complexity. 
		
		As reported in Table~\ref{t:hyper-parameter}, the extra computational complexity in our proposed methods is acceptable in practice. 
		Moreover, in \jname, we make full use of label information to generate positive semantic representations for input sentences and use in-batch negative sampling to obtain the negative semantic representations for each input sentence. 
		Furthermore, to alleviate the computational complexity of pair-wise contrastive learning, we leverage the same method as~\cite{gao2021simcse}, which transfers the pair-wise calculation into a cross-entropy calculation during implementation. This operation can largely decrease the computational complexity of pair-wise contrastive learning.

		\subsection{Overall Experimental Results}
		In this section, we will give a detailed analysis of experimental results by reporting accuracy and error rate comparison with backbones. 
		Note that error rate comparison is based on corresponding backbone. 
		For example, \cname-BERT-(base) achieves $+8.25\%$ improvement in Table~\ref{t:nli-result}(16) denotes that it decreases the error rate by $8.25\%$ compared with backbone BERT-(base). 
		Detailed analysis is reported in following parts.
		
		\subsubsection{\textbf{Performance on NLI task}}
		Table~\ref{t:nli-result} reports the results on NLI task. 
		We can conclude that our proposed \cname~and \jname~achieve highly comparable performance over all datasets. Moreover, \jname-RoBERTa-(base) has the best performance. 
		To achieve such advantages, we first select PLMs as backbones, so that knowledge from large corpora can be accessed. This is one of the reasons that our methods outperform other baselines, especially better than these PLMs-free baselines (Table~\ref{t:nli-result}(1)-(4)) by a large margin. 
		Second, we develop a novel R$^2$ task to help models fully exploit label information in a one-hot manner. 
		Along this line, \cname~and \jname~can obtain intra-class and inter-class knowledge among input texts with the same or different labels, which is in favor of achieving better performance than all baselines, including PLMs baselines (Table~\ref{t:nli-result}(5)-(9)). 
		Moreover, we design to incorporate fine-grained descriptions from WordNet into label embedding learning, so that particular label semantics can be better represented. 
		This is another vital reason that \jname~achieves the best performance, which also demonstrates the effectiveness of our proposed description-enhanced label embedding method.

		Moreover, incorporating label information (one-hot encoding or label embedding) can better ensure model performance when dealing with difficult examples (Hard test in Table~\ref{t:nli-result}), in which text sequences with obviously identical words have been removed~\cite{gururangan2018annotation}. 
		By employing R$^2$ task to measure labels in a one-hot manner, \cname~can better exploit implicit information from input text, which leads to an inspiring performance. 
		Moreover, label embedding methods can exploit label information better than one-hot encoding methods. 
		Therefore, we observe the superiority of label embedding methods. 
		Furthermore, \jname~employs fine-grained descriptions to enhance label embedding learning. 
		Thus, it further improves the performance of label embedding methods and achieves the best performance.
		
		To make a fair comparison, we replace basic encoders of label embedding methods~(e.g., LEAM, EXAM) with the same PLMs backbones as our methods did, and report results in Table~\ref{t:nli-result}(10)-(15), in which \jname~still has better performance. 
		The advantages of \jname~lie in the more advanced label encoder and the consideration of denoising operation for additional knowledge. 
		For label encoder, \jname~not only learns the embedding during the model training so that the learned results are suitable for the current task, but also employs multi-aspect real sentences from WordNet as descriptions to enhance label semantic modelling.
		Meanwhile, considering that introducing fine-grained descriptions will import unexpected noise, we develop a novel mutual interaction module based on CL to leverage attention mechanism to select the most relevant parts from input text and labels simultaneously. 
		Then, with the help of CL framework, \jname~is able to use proper fine-grained knowledge to enhance the understanding of labels, and improve the model performance. 
		
		\begin{figure}
			\centering
			\includegraphics[width=0.4\textwidth]{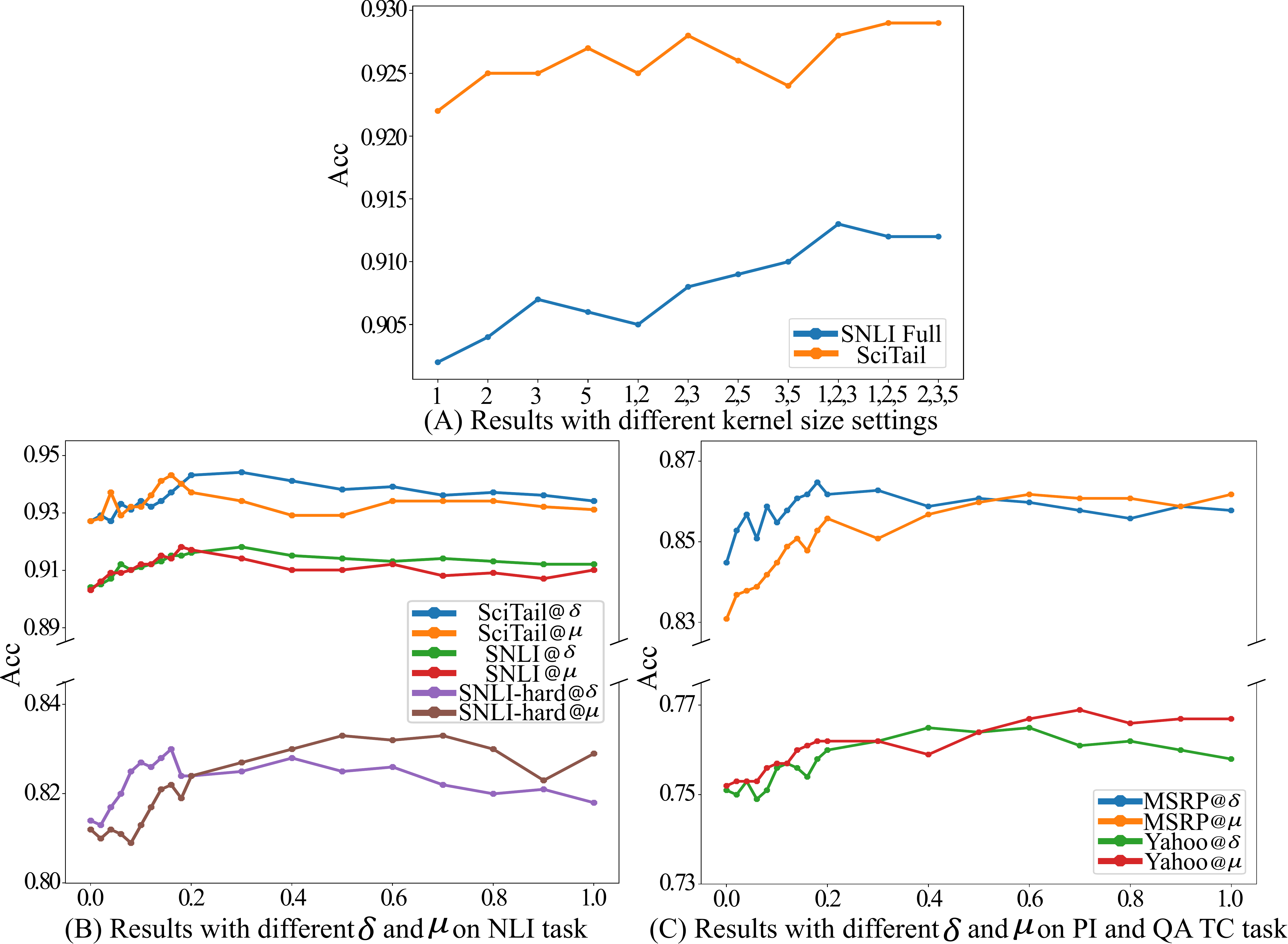}
			\caption{Results of \cname~with different kernel sizes and \jname~with different $\sigma$ and $\mu$ on different datasets.}
			\label{f:lambda_sensitive}
		\end{figure}

		\subsubsection{\textbf{Performance on PI task}}
		Apart from NLI task, we also select PI task to evaluate the model performance. PI task concerns whether two sentences express the same meaning and has broad applications in question answering communities\footnote{https://www.quora.com/}. 
		Table~\ref{t:pi-qa-result} reports corresponding results. 
		We observe that \cname~and \jname~still achieve highly competitive performance over other baselines. 
		For one thing, the results demonstrate that though PI task is a binary classification task, its labels still contain some useful semantics. 
		For another, we can conclude that our proposed R$^2$ task and label embedding methods with fine-grained descriptions are useful for exploiting label information and boosting model performance. 
		
		Besides, we obtain that almost all models have a better performance on Quora than MSRP. One possible reason is that Quora has more data~(over 400k sentence pairs in Quora v.s. 5,801 sentence pairs in MSRP). 
		Besides, inter-sentence interaction is probably another reason. 
		Lan \textit{et al.}~\cite{lan2018neural} has observed that Quora dataset contains many sentence pairs with less complicated interactions~(many identical words in sentence pairs). 
		Therefore, we can obtain that ALBERT achieves the best performance on MSRP dataset.
		
		\begin{table}
			\centering
			\caption{Ablation performance (accuracy) of \cname-BERT-(base).}
			\begin{footnotesize}
				\begin{tabular}{lcc} \hline
					\textbf{Model} & \textbf{SNLI Full Test} & \textbf{SciTail Test} \\ \hline
					(1)~BERT-(base) & 90.3\% & 92.0\%  \\ \hline
					(2)~\cname~(w/o local encoder) & 90.7\% & 92.6\%  \\ 
					(3)~\cname~(w/o R$^2$ task learning) & 90.5\% & 92.3\%  \\ 
					(4)~\cname~(w/o triplet loss) & 90.9\% & 92.6\%  \\
					(5)~\cname~(w/ NT-Xent loss) & \textbf{91.3}\% & \textbf{93.3}\%  \\ \hline
					(6)~\cname-BERT-(base)~ & \underline{91.1}\%  & \underline{92.9}\% \\ 
					\hline
				\end{tabular}
			\end{footnotesize}
			\label{t:r2-ablation}
		\end{table}
		
		\subsubsection{\textbf{Performance on Sentiment Analysis and QA Topic Classification}}
		\label{s:qa_analysis}
		To make a better evaluate, we further conduct experiments on \textit{SST-5} and \textit{Yahoo! Answer} datasets, which have more complex and realistic labels (e.g., \textit{somewhat positive, Business}). 
		Table~\ref{t:pi-qa-result} summarizes the results. 
		Similarly, our proposed methods achieve stable and comparable performance. 
		Moreover, Different from previous experiments, label embedding baselines~(i.e., EXAM, LGDSC) do not achieve a better improvement, compared with PLMs baselines. 
		One possible reason is that embedding these complicated labels with coarse-grained information cannot capture sufficient useful information for label exploitation and text classification. 
		
		Moreover, compared with PLMs baselines, the improvement of \jname~is not so obvious. 
		After analyzing the dataset deeply, we observe that some QA pairs in this dataset have more than 512 words, as well as irregular textual expressions. 
		These input noises will do harm to label utilization since we leverage text representations to guide the denoising process of label description utilization. 
		Meanwhile, we only select no more than 3 descriptions manually, which may be insufficient for those labels that have a wealth of semantics.
		Furthermore, the particular meaning of label words requires a precise selection of label descriptions. 
		However, \jname~leverages attention mechanism to select the most relevant parts from label embedding $\bm{E}$, where vanilla embedding and description embedding have already been integrated. 
		Applying denoising operation at an earlier stage may achieve better performance. 
		Nevertheless, on the contrary, \jname~has made an early attempt at leveraging fine-grained knowledge to enhance label embedding and denoising of knowledge utilization, which also illustrates the advancement of \jname.
		
		\begin{figure*}
			\centering
			\includegraphics[width=0.94\textwidth]{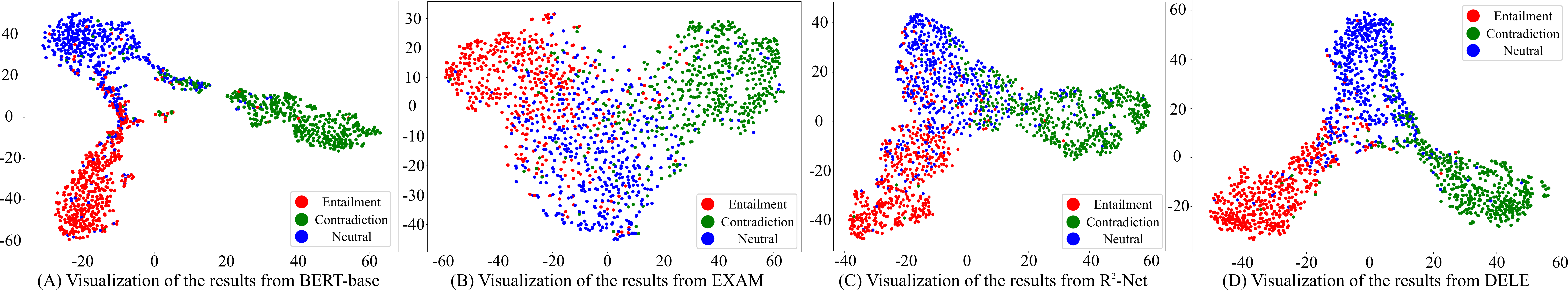}
			\caption{Visualization of semantic representations from two representative baselines~(BERT-(base)(A) and EXAM (B)), as well as our proposed \cname~(C) and \jname~(D). Note that BERT-(base) is selected as the backbone in this experiment.} 
			\label{f:case}
		\end{figure*}
		
		\subsubsection{\textbf{Parameter Sensitive Test}}
		As illustrated in Eq.~\ref{eq:local-encoding} and Eq.~\ref{eq:jloss}, kernel sizes $d_k$ in \cname, and $\{\delta, \mu\}$ in \jname~are essential for model performance.
		To this end, we conduct additional parameter sensitive test to verify their impact. 
		
		For kernel sizes $d_k$, it aims to extract local information of input text at different scales. 
		Results with different kernel size settings are reported in Fig.~\ref{f:lambda_sensitive}(A). 
		We can observe that \cname~will have better performance with more kernel sizes, in which $\{1, 2, 3\}$, $\{1, 2, 5\}$, and $\{2, 3, 5\}$ achieve the best performance. 
		Moreover, using $\{1, 2, 3\}$ as kernel sizes have more stable performance. We speculate the possible reason is that these kernel sizes can pay more attention to uni-gram, bi-gram, as well as tri-gram information, which is helpful for enhancing the representation learning of PLMs. 
		Meanwhile, using $\{2, 3, 5\}$ as kernel sizes has similar performance, which also demonstrates its effectiveness.
		
		For $\sigma$ and $\mu$, when removing CL or R$^2$ task~(i.e., $\delta=0$ or $\mu=0$), model performance cannot be comparable with the performance with CL or R$^2$ task (i.e., $\delta \neq 0$ or $\mu \neq 0$), demonstrating the necessity of these two SSL frameworks.  
		Moreover, with the increasing of label size or label semantics, the best values of $\delta$ and $\mu$ also become larger. 
		For one thing, this phenomenon proves that our proposed SSL can extract more critical information from more diverse and realistic labels, and have a bigger impact on model performance,  
		proving the necessity of better label utilization methods. 
		Furthermore, we observe that $\mu$ has a bigger impact than $\alpha$ when labels are harder, which is somewhat counterintuitive and different from other results. 
		The possible reason is the underutilization of label information. 
		Limited information and unexpected noise will decrease the effectiveness of our proposed CL framework. 
		This finding is similar to the observation in Section~\ref{s:qa_analysis}.

		\begin{table}
			\centering
			\caption{Ablation performance (accuracy) of \jname-BERT-(base).}
			\begin{tabular}{lcc} \hline
				\textbf{Model} & \textbf{SNLI Full Test} & \textbf{SciTail Test} \\ \hline
				(1)~BERT-(base) & 90.3\% & 93.1\% \\
				(2)~BERT + Atten & 90.6\% & 93.5\% \\ \hline
				(3)~$\bar{\bm{h}}_s$(w/o mutual interaction) & 90.4\% & 93.2\%\\
				(4)~$\bar{\bm{h}}_e$(w/o mutual interaction) & \textit{43.7\%} & \textit{65.4\%}\\ 
				
				(5)~$\bar{\bm{h}}_s$(w/o description) & \underline{90.8}\% & \underline{93.8}\%\\ 
				(6)~$\bar{\bm{h}}_e$(w/o description) & 89.3\% & 90.1\%\\
				
				(7)~$\bar{\bm{h}}_s$(w/o R$^2$ task)) & 90.3\% & 92.9\%\\ 
				(8)~$\bar{\bm{h}}_e$(w/o R$^2$ task) & 88.5\% & 90.4\%\\
				\hline
				
				(9)~\jname-BERT-(base) & \textbf{91.3}\% & \textbf{94.1}\%\\ 
				\hline
			\end{tabular}
			\label{t:dlan-ablation}
		\end{table}
		
		\subsubsection{\textbf{Ablation Study}}
		The overall experiments have proved the superiority of our proposed methods. 
		However, which parts in \cname~and \jname~play a more important role in label utilization and performance improvement is still unclear.  
		Therefore, we perform an ablation study to conduct a comprehensive analysis. Corresponding results are reported in Table~\ref{t:r2-ablation} and Table~\ref{t:dlan-ablation}. 
		Note that we select BERT-(base) as the backbone to compare the importance of each part. Leveraging RoBERTa-(base) as the backbone will obtain similar results. 
		
		(a) For \cname, we conduct experiments on \textit{CNN-based local encoder}, \textit{R$^2$ task classification}, as well as \textit{loss function}. 
		Results are summarized in Table~\ref{t:r2-ablation}. 
		According to the results, we can observe varying degrees of model performance decline. Among all of them, R$^2$ task has the biggest impact, and triple loss has a relatively small impact on the model performance. 
		Replacing triplet loss with contrastive loss will improve model performance slightly. 
		These observations prove that  R$^2$ task is more important for relation information utilization. 
		R$^2$ task and better contrastive loss will further improve model performance.
		
		(b) For \jname, we focus on \textit{label embedding module}, \textit{mutual interaction module}, and \textit{R$^2$ task}. Results are reported in Table~\ref{t:dlan-ablation}. 
		Here, ``BERT+Atten'' denotes leveraging vanilla label embedding to select important parts of text without CL
		``$\bar{\bm{h}}_s$ (w/o mutual interaction)'' indicates leveraging $\bar{\bm{h}}_s$ for classification and no interactions between input text and labels. 
		Other ablation experiments have similar settings. 
		According to the results, we obtain some observations. 
		First of all, results (3-4) demonstrate the necessity of mutual interactions between input text and labels, especially result (4). When there is no interaction between input text and labels, label semantics cannot handle final classification at all.
		Second, results (5-6) prove that utilizing fine-grained descriptions is helpful for semantic understanding and performance improvement. 
		Moreover, results (7-8) inspire us that R$^2$ task can provide additional signals for label utilization, which demonstrates the effectiveness and necessity of better label utilization methods.
		Last but not least, the comparison between $\bar{\bm{h}}_s$ and $\bar{\bm{h}}_e$ with the same setting (e.g., w/o description) indicates that CL can constrain learned representations from different views to have similar semantics. However, $\bar{\bm{h}}_e$ cannot be the replacement for input text, only an enhancement to input text.

		\subsubsection{\textbf{Case Study of \cname~and \jname}}
		To provide some intuitionistic examples for explaining the superiority of our models, 
		we sample 1,500 sentence pairs from SNLI Test and send them to BERT-(base) baseline, EXAM-BERT-(base) baseline, \cname, and \jname~to generate representations.
		Then, we use t-sne~\cite{maaten2008visualizing} to visualize them with \textit{the same parameter settings}.
		Fig.~\ref{f:case} reports corresponding results.  
		By comparing these four figures, we observe that representations generated by \jname~have closer intra-class distances and more distinguishable inter-class distances. 
		In other words, by considering fine-grained knowledge and mutual interactions, \jname~is able to pull together samples with the same label and push apart samples with different labels, which is helpful for better text classification. 
		These observations not only explain why \jname~achieves impressive performance, but also demonstrate the necessity of more detailed and cleaner label descriptions, as well as more comprehensive analysis between input text and labels. 
		All of these are very helpful for label semantics utilization and text classification performance improvement.

		\section{Discussion}
		\label{s:discussion}
		Here, we discuss the impact of the future directions of this study. 
		To fully exploit label information for text classification, we develop a novel \cname~and \jname~from the one-hot encoding perspective and label embedding perspective separately. 
		Despite the progress we have achieved, our work can also provide some inspiration for relative research. 
		For example, in Extreme Multi-label classification tasks (XMC), label embedding matrix learning is a challenging problem. Inappropriate embedding methods may perform worse than sparse one-vs-all and partitioning approaches in XMC. To this end, our proposed methods provide a novel strategy to generate accurate label embeddings, which may inspire relative research, such as using additional descriptions to distinguish the similarity relations among hundreds of labels. 
		
		Meanwhile, our proposed methods still have space for further improvement. 
		For example, when selecting fine-grained descriptions as additional knowledge, we employ a manual approach to select relevant descriptions, which is insufficient for complex labels and impractical for extreme multi-label classification. 
		Therefore, better knowledge selection methods or multi-modal knowledge (e.g., relations and hierarchical structures among different labels) are needed to boost semantic representation learning further. 
		When talking about additional information, we only consider label descriptions. The relations and hierarchical structures among labels are also beneficial for label semantic modelling. 
		Thus, in the future, we plan to exploit the label relations in XMC and use Graph Neural Network~(GNN) to obtain relation-aware label embeddings. Then, we can transfer our proposed methods to XMC by incorporating relation-aware label embeddings with our proposed description-enhanced label embeddings for model performance enhancement.

		\section{Conclusion and Future Work}
		\label{s:conclusion}
		In this paper, we argued that current text classification methods are deficient in label utilization and ignore the guidance and semantic information contained in labels. 
		Then, we presented a study on the exploitation of labels from a one-hot encoding manner and label embedding manners. 
		Specifically, inspired by self-supervised learning methods used in PLMs, we developed a novel Relation of Relation (R$^2$) task to mine the potential of labels from a one-hot encoding perspective. 
		Based on this novel self-supervised task, we designed a simple but effective method named \cname~for text classification. 
		Meanwhile, triplet loss is employed to constrain \cname~for better label relation analysis. 
		One step further, since one-hot encoding method still has some weaknesses in exploiting label information, we focused on label embedding methods and proposed a novel \jname~for better label utilization. 
		In \jname, fine-grained descriptions from WordNet are adopted to complete the label embedding learning. And a mutual interaction module is designed to denoise the additional knowledge and select the most relevant parts from input text and labels simultaneously. 
		Finally, extensive experiments on multiple benchmark datasets demonstrate the superiority of \cname~and \jname, as well as the usefulness of initial attempts on fine-grained knowledge utilization for label embedding.

		
		\section*{Acknowledgment}

		This research was partially supported by grants from the Young Scientists Fund of the National Natural Science Foundation of China (No. 62006066), the National Natural Science Foundation of China (No. 61727809, 61922073, and 72188101), the Fundamental Research Funds for the Central Universities (JZ2021HGTB0075), joint Funds of the National Natural Science Foundation of China (U22A2094), the Open Project Program of the National Laboratory of Pattern Recognition.

		\ifCLASSOPTIONcaptionsoff
		\newpage
		
		\fi

		
		
		%
		\bibliographystyle{IEEEtran}
		\bibliography{10_myReference}

\begin{thebibliography}{10}
\providecommand{\url}[1]{#1}
\csname url@samestyle\endcsname
\providecommand{\newblock}{\relax}
\providecommand{\bibinfo}[2]{#2}
\providecommand{\BIBentrySTDinterwordspacing}{\spaceskip=0pt\relax}
\providecommand{\BIBentryALTinterwordstretchfactor}{4}
\providecommand{\BIBentryALTinterwordspacing}{\spaceskip=\fontdimen2\font plus
\BIBentryALTinterwordstretchfactor\fontdimen3\font minus
  \fontdimen4\font\relax}
\providecommand{\BIBforeignlanguage}[2]{{%
\expandafter\ifx\csname l@#1\endcsname\relax
\typeout{** WARNING: IEEEtran.bst: No hyphenation pattern has been}%
\typeout{** loaded for the language `#1'. Using the pattern for}%
\typeout{** the default language instead.}%
\else
\language=\csname l@#1\endcsname
\fi
#2}}
\providecommand{\BIBdecl}{\relax}
\BIBdecl

\bibitem{dolan2005automatically}
W.~B. Dolan and C.~Brockett, ``Automatically constructing a corpus of
  sentential paraphrases,'' in \emph{IWP}, 2005.

\bibitem{Kim2018SemanticSM}
S.~Kim, J.-H. Hong, I.~Kang, and N.~Kwak, ``Semantic sentence matching with
  densely-connected recurrent and co-attentive information,'' in \emph{AAAI},
  2019, pp. 6586--6593.

\bibitem{chen2017reading}
D.~Chen, A.~Fisch, J.~Weston, and A.~Bordes, ``Reading wikipedia to answer
  open-domain questions,'' in \emph{ACL}, 2017, pp. 1870--1879.

\bibitem{yilmaz2023multi}
S.~F. Yilmaz, E.~B. Kaynak, A.~Koç, H.~Dibeklioğlu, and S.~S. Kozat,
  ``Multi-label sentiment analysis on 100 languages with dynamic weighting for
  label imbalance,'' \emph{IEEE TNNLS}, vol.~34, no.~1, pp. 331--343, 2023.

\bibitem{huang2022attention}
F.~Huang, X.~Li, C.~Yuan, S.~Zhang, J.~Zhang, and S.~Qiao,
  ``Attention-emotion-enhanced convolutional lstm for sentiment analysis,''
  \emph{IEEE TNNLS}, vol.~33, no.~9, pp. 4332--4345, 2022.

\bibitem{zhu2021senti}
L.~Zhu, W.~Li, Y.~Shi, and K.~Guo, ``Sentivec: Learning sentiment-context
  vector via kernel optimization function for sentiment analysis,'' \emph{IEEE
  TNNLS}, vol.~32, no.~6, pp. 2561--2572, 2021.

\bibitem{Liu2010LearningTR}
T.-Y. Liu, ``Learning to rank for information retrieval,'' in \emph{Found.
  Trends Inf. Retr.}, vol.~3, 2009, pp. 225--331.

\bibitem{liu2018finding}
Q.~Liu, Z.~Huang, Z.~Huang, C.~Liu, E.~Chen, Y.~Su, and G.~Hu, ``Finding
  similar exercises in online education systems,'' in \emph{SIGKDD}, 2018, pp.
  1821--1830.

\bibitem{serban2016building}
I.~V. Serban, A.~Sordoni, Y.~Bengio, A.~C. Courville, and J.~Pineau, ``Building
  end-to-end dialogue systems using generative hierarchical neural network
  models,'' in \emph{AAAI}, 2016, pp. 3776--3783.

\bibitem{zhang2019drr}
K.~Zhang, G.~Lv, L.~Wang, L.~Wu, E.~Chen, F.~Wu, and X.~Xie, ``Drr-net: Dynamic
  re-read network for sentence semantic matching,'' in \emph{AAAI}, vol.~33,
  2019, pp. 7442--7449.

\bibitem{zhang2021ladra}
K.~Zhang, G.~Lv, L.~Wu, E.~Chen, Q.~Liu, and M.~Wang, ``Ladra-net: Locally
  aware dynamic reread attention net for sentence semantic matching,''
  \emph{IEEE TNNLS}, pp. 1--14, 2021.

\bibitem{tan2022dynamic}
Z.~Tan, J.~Chen, Q.~Kang, M.~Zhou, A.~Abusorrah, and K.~Sedraoui, ``Dynamic
  embedding projection-gated convolutional neural networks for text
  classification,'' \emph{IEEE TNNLS}, vol.~33, no.~3, pp. 973--982, 2022.

\bibitem{zhang2018ImageEnhance}
Z.~Kun, L.~Guangyi, W.~Le, C.~Enhong, L.~Qi, and W.~Han, ``Image-enhanced
  multi-level sentence representation net for natural language inference,'' in
  \emph{IEEE ICDM}, 2018, pp. 747--756.

\bibitem{guo2020label}
B.~Guo, S.~Han, X.~Han, H.~Huang, and T.~Lu, ``Label confusion learning to
  enhance text classification models,'' \emph{Arxiv}, vol. abs/2012.04987,
  2020.

\bibitem{xiong2021fusing}
Y.~Xiong, Y.~Feng, H.~Wu, H.~Kamigaito, and M.~Okumura, ``Fusing label
  embedding into {BERT}: An efficient improvement for text classification,'' in
  \emph{Findings of ACL-IJCNLP}, 2021, pp. 1743--1750.

\bibitem{zhang-etal-2018-multi}
H.~Zhang, L.~Xiao, W.~Chen, Y.~Wang, and Y.~Jin, ``Multi-task label embedding
  for text classification,'' in \emph{EMNLP}, 2018, pp. 4545--4553.

\bibitem{gururangan2018annotation}
S.~Gururangan, S.~Swayamdipta, O.~Levy, R.~Schwartz, S.~R. Bowman, and N.~A.
  Smith, ``Annotation artifacts in natural language inference data,'' in
  \emph{NAACL-HLT}, 2018, pp. 107--112.

\bibitem{xiao2019label}
L.~Xiao, X.~Huang, B.~Chen, and L.~Jing, ``Label-specific document
  representation for multi-label text classification,'' in \emph{EMNLP-IJCNLP},
  2019, pp. 466--475.

\bibitem{zhang2022long}
R.~Zhang, Y.-S. Wang, Y.~Yang, D.~Yu, T.~Vu, and L.~Lei, ``Long-tailed extreme
  multi-label text classification with generated pseudo label descriptions,''
  \emph{Arxiv}, vol. abs/2204.00958, 2022.

\bibitem{zhang2021making}
K.~Zhang, L.~Wu, G.~Lv, M.~Wang, E.~Chen, and S.~Ruan, ``Making the relation
  matters: Relation of relation learning network for sentence semantic
  matching,'' in \emph{AAAI}, 2021, pp. 14\,411--14\,419.

\bibitem{devlin2018bert}
J.~Devlin, M.-W. Chang, K.~Lee, and K.~Toutanova, ``Bert: Pre-training of deep
  bidirectional transformers for language understanding,'' \emph{Arxiv}, vol.
  abs/1810.04805, 2018.

\bibitem{kim2014convolutional}
Y.~Kim, ``Convolutional neural networks for sentence classification,''
  \emph{Arxiv}, vol. abs/1408.5882, 2014.

\bibitem{Chung2014EmpiricalEO}
J.~Chung, C.~Gulcehre, K.~Cho, and Y.~Bengio, ``Empirical evaluation of gated
  recurrent neural networks on sequence modeling,'' \emph{Arxiv}, vol.
  abs/1412.3555, 2014.

\bibitem{Parikh2016ADA}
A.~P. Parikh, O.~T{\"a}ckstr{\"o}m, D.~Das, and J.~Uszkoreit, ``A decomposable
  attention model for natural language inference,'' in \emph{EMNLP}, 2016, pp.
  2249--2255.

\bibitem{bowman2015large}
S.~R. Bowman, G.~Angeli, C.~Potts, and C.~D. Manning, ``A large annotated
  corpus for learning natural language inference,'' in \emph{EMNLP}, 2015, p.
  632–642.

\bibitem{iyer2017first}
S.~Iyer, N.~Dandekar, and K.~Csernai, ``First quora dataset release: Question
  pairs,'' 2017.

\bibitem{socher2013recursive}
R.~Socher, A.~Perelygin, J.~Wu, J.~Chuang, C.~D. Manning, A.~Y. Ng, and
  C.~Potts, ``Recursive deep models for semantic compositionality over a
  sentiment treebank,'' in \emph{EMNLP}, 2013, pp. 1631--1642.

\bibitem{Liu2019RoBERTaAR}
Y.~Liu, M.~Ott, N.~Goyal, J.~Du, M.~Joshi, D.~Chen, O.~Levy, M.~Lewis,
  L.~Zettlemoyer, and V.~Stoyanov, ``Roberta: A robustly optimized bert
  pretraining approach,'' \emph{Arxiv}, vol. abs/1907.11692, 2019.

\bibitem{xu2020semi}
W.~Xu and Y.~Tan, ``Semisupervised text classification by variational
  autoencoder,'' \emph{IEEE TNNLS}, vol.~31, no.~1, pp. 295--308, 2020.

\bibitem{zeng2014relation}
D.~Zeng, K.~Liu, S.~Lai, G.~Zhou, and J.~Zhao, ``Relation classification via
  convolutional deep neural network,'' in \emph{COLING}, 2014, pp. 2335--2344.

\bibitem{liu2020finding}
X.~Liu, L.~Mou, H.~Cui, Z.~Lu, and S.~Song, ``Finding decision jumps in text
  classification,'' \emph{Neurocomputing}, vol. 371, pp. 177--187, 2020.

\bibitem{cai2020hybrid}
L.~Cai, Y.~Song, T.~Liu, and K.~Zhang, ``A hybrid bert model that incorporates
  label semantics via adjustive attention for multi-label text
  classification,'' \emph{IEEE Access}, vol.~8, pp. 152\,183--152\,192, 2020.

\bibitem{mueller2022label}
A.~Mueller, J.~Krone, S.~Romeo, S.~Mansour, E.~Mansimov, Y.~Zhang, and D.~Roth,
  ``Label semantic aware pre-training for few-shot text classification,''
  \emph{arXiv preprint arXiv:2204.07128}, 2022.

\bibitem{liu2022co}
M.~Liu, L.~Liu, J.~Cao, and Q.~Du, ``Co-attention network with label embedding
  for text classification,'' \emph{Neurocomputing}, vol. 471, pp. 61--69, 2022.

\bibitem{zhu2022generating}
X.~Zhu, Z.~Peng, J.~Guo, and S.~Dietze, ``Generating effective label
  description for label-aware sentiment classification,'' \emph{Expert Systems
  with Applications}, p. 119194, 2022.

\bibitem{du2019explicit}
C.~Du, Z.~Chen, F.~Feng, L.~Zhu, T.~Gan, and L.~Nie, ``Explicit interaction
  model towards text classification,'' in \emph{AAAI}, vol.~33, 2019, pp.
  6359--6366.

\bibitem{rivas2020efficient}
K.~Rivas~Rojas, G.~Bustamante, A.~Oncevay, and M.~A. Sobrevilla~Cabezudo,
  ``Efficient strategies for hierarchical text classification: External
  knowledge and auxiliary tasks,'' in \emph{ACL}, 2020, pp. 2252--2257.

\bibitem{wu2022effective}
H.~Wu, S.~Qin, R.~Nie, J.~Cao, and S.~Gorbachev, ``Effective collaborative
  representation learning for multilabel text categorization,'' \emph{IEEE
  TNNLS}, vol.~33, no.~10, pp. 5200--5214, 2022.

\bibitem{wang2021concept}
X.~Wang, L.~Zhao, B.~Liu, T.~Chen, F.~Zhang, and D.~Wang, ``Concept-based label
  embedding via dynamic routing for hierarchical text classification,'' in
  \emph{ACL-IJCNLP}, 2021, pp. 5010--5019.

\bibitem{wang2020contrastive}
N.~Wang, W.~Zhou, and H.~Li, ``Contrastive transformation for self-supervised
  correspondence learning,'' in \emph{AAAI}, 2021, pp. 10\,174--10\,182.

\bibitem{wu2020clear}
Z.~Wu, S.~Wang, J.~Gu, M.~Khabsa, F.~Sun, and H.~Ma, ``Clear: Contrastive
  learning for sentence representation,'' \emph{Arxiv}, vol. abs/2012.15466,
  2020.

\bibitem{khosla2020supervised}
P.~Khosla, P.~Teterwak, C.~Wang, A.~Sarna, Y.~Tian, P.~Isola, A.~Maschinot,
  C.~Liu, and D.~Krishnan, ``Supervised contrastive learning,'' in
  \emph{NeurIPS}, 2020, pp. 1--13.

\bibitem{cai2020negatives}
T.~T. Cai, J.~Frankle, D.~J. Schwab, and A.~S. Morcos, ``Are all negatives
  created equal in contrastive instance discrimination?'' \emph{Arxiv}, vol.
  abs/2010.06682, 2020.

\bibitem{Chen2020ASF}
T.~Chen, S.~Kornblith, M.~Norouzi, and G.~E. Hinton, ``A simple framework for
  contrastive learning of visual representations,'' \emph{Arxiv}, vol.
  abs/2002.05709, 2020.

\bibitem{he2020momentum}
K.~He, H.~Fan, Y.~Wu, S.~Xie, and R.~Girshick, ``Momentum contrast for
  unsupervised visual representation learning,'' \emph{Arxiv}, vol.
  abs/1911.05722, 2020.

\bibitem{he2021masked}
K.~He, X.~Chen, S.~Xie, Y.~Li, P.~Dollár, and R.~Girshick, ``Masked
  autoencoders are scalable vision learners,'' \emph{Arxiv}, vol.
  abs/2111.06377, 2021.

\bibitem{gao2021simcse}
T.~Gao, X.~Yao, and D.~Chen, ``Simcse: Simple contrastive learning of sentence
  embeddings,'' in \emph{EMNLP}, 2021, pp. 6894--6910.

\bibitem{li2021contrastive}
Y.~Li, P.~Hu, Z.~Liu, D.~Peng, J.~T. Zhou, and X.~Peng, ``Contrastive
  clustering,'' in \emph{AAAI}, 2021, pp. 8547--8555.

\bibitem{yang2021partially}
M.~Yang, Y.~Li, Z.~Huang, Z.~Liu, P.~Hu, and X.~Peng, ``Partially view-aligned
  representation learning with noise-robust contrastive loss,'' in \emph{CVPR},
  2021, pp. 1134--1143.

\bibitem{yang2022robust}
M.~Yang, Y.~Li, P.~Hu, J.~Bai, J.~C. Lv, and X.~Peng, ``Robust multi-view
  clustering with incomplete information,'' \emph{IEEE TPAMI}, 2022.

\bibitem{Peters2018DeepCW}
M.~E. Peters, M.~Neumann, M.~Iyyer, M.~Gardner, C.~Clark, K.~Lee, and
  L.~Zettlemoyer, ``Deep contextualized word representations,'' \emph{Arxiv},
  vol. abs/1802.05365, 2018.

\bibitem{sennrich2015neural}
R.~Sennrich, B.~Haddow, and A.~Birch, ``Neural machine translation of rare
  words with subword units,'' in \emph{ACL}, 2016, p. 1715–1725.

\bibitem{Chen-Qian2017ACL}
Q.~Chen, X.~Zhu, Z.~Ling, S.~Wei, H.~Jiang, and D.~Inkpen, ``Enhanced lstm for
  natural language inference,'' in \emph{ACL}, 2017, pp. 1657--1668.

\bibitem{zhang2017context}
K.~Zhang, E.~Chen, Q.~Liu, C.~Liu, and G.~Lv, ``A context-enriched neural
  network method for recognizing lexical entailment.'' in \emph{AAAI}, 2017,
  pp. 3127--3133.

\bibitem{mou2016natural}
L.~Mou, R.~Men, G.~Li, Y.~Xu, L.~Zhang, R.~Yan, and Z.~Jin, ``Natural language
  inference by tree-based convolution and heuristic matching,'' in \emph{ACL},
  2016, pp. 130--136.

\bibitem{weeds2014learning}
J.~Weeds, D.~Clarke, J.~Reffin, D.~Weir, and B.~Keller, ``Learning to
  distinguish hypernyms and co-hyponyms,'' in \emph{COLING}, 2014, pp.
  2249--2259.

\bibitem{Schroff2015FaceNetAU}
F.~Schroff, D.~Kalenichenko, and J.~Philbin, ``Facenet: A unified embedding for
  face recognition and clustering,'' in \emph{CVPR}, 2015, pp. 815--823.

\bibitem{chen2020simple}
T.~Chen, S.~Kornblith, M.~Norouzi, and G.~Hinton, ``A simple framework for
  contrastive learning of visual representations,'' in \emph{ICML}, 2020, pp.
  1597--1607.

\bibitem{tay2018co}
Y.~Tay, A.~T. Luu, and S.~C. Hui, ``Co-stack residual affinity networks with
  multi-level attention refinement for matching text sequences,'' in
  \emph{ACL}, 2018, pp. 4492--4502.

\bibitem{yang2019simple}
R.~Yang, J.~Zhang, X.~Gao, F.~Ji, and H.~Chen, ``Simple and effective text
  matching with richer alignment features,'' in \emph{ACL}, 2019, pp.
  4699--4709.

\bibitem{Lan2020ALBERTAL}
Z.~Lan, M.~Chen, S.~Goodman, K.~Gimpel, P.~Sharma, and R.~Soricut, ``Albert: A
  lite bert for self-supervised learning of language representations,'' in
  \emph{ICLR}, 2020, pp. 1--13.

\bibitem{wang2018jointembedding}
G.~Wang, C.~Li, W.~Wang, Y.~Zhang, D.~Shen, X.~Zhang, R.~Henao, and L.~Carin,
  ``Joint embedding of words and labels for text classification,'' in
  \emph{ACL}, 2018, pp. 2321--2331.

\bibitem{marelli2014semeval}
M.~Marelli, L.~Bentivogli, M.~Baroni, R.~Bernardi, S.~Menini, and
  R.~Zamparelli, ``Semeval-2014 task 1: Evaluation of compositional
  distributional semantic models on full sentences through semantic relatedness
  and textual entailment,'' in \emph{SemEval}, 2014, pp. 1--8.

\bibitem{khot2018scitail}
T.~Khot, A.~Sabharwal, and P.~Clark, ``Scitail: A textual entailment dataset
  from science question answering,'' in \emph{AAAI}, 2018, pp. 5189--5197.

\bibitem{howard2018universal}
J.~Howard and S.~Ruder, ``Universal language model fine-tuning for text
  classification,'' in \emph{ACL}, 2018, p. 328–339.

\bibitem{lan2018neural}
W.~Lan and W.~Xu, ``Neural network models for paraphrase identification,
  semantic textual similarity, natural language inference, and question
  answering,'' in \emph{COLING}, 2018, pp. 3890--3902.

\bibitem{maaten2008visualizing}
L.~v.~d. Maaten and G.~Hinton, ``Visualizing data using t-sne,'' \emph{Journal
  of machine learning research}, vol.~9, no. Nov, pp. 2579--2605, 2008.

\end{thebibliography}
		
		\vspace{-16mm}
		\begin{IEEEbiography}[{\includegraphics[width=1in,height=1.25in,clip,keepaspectratio]{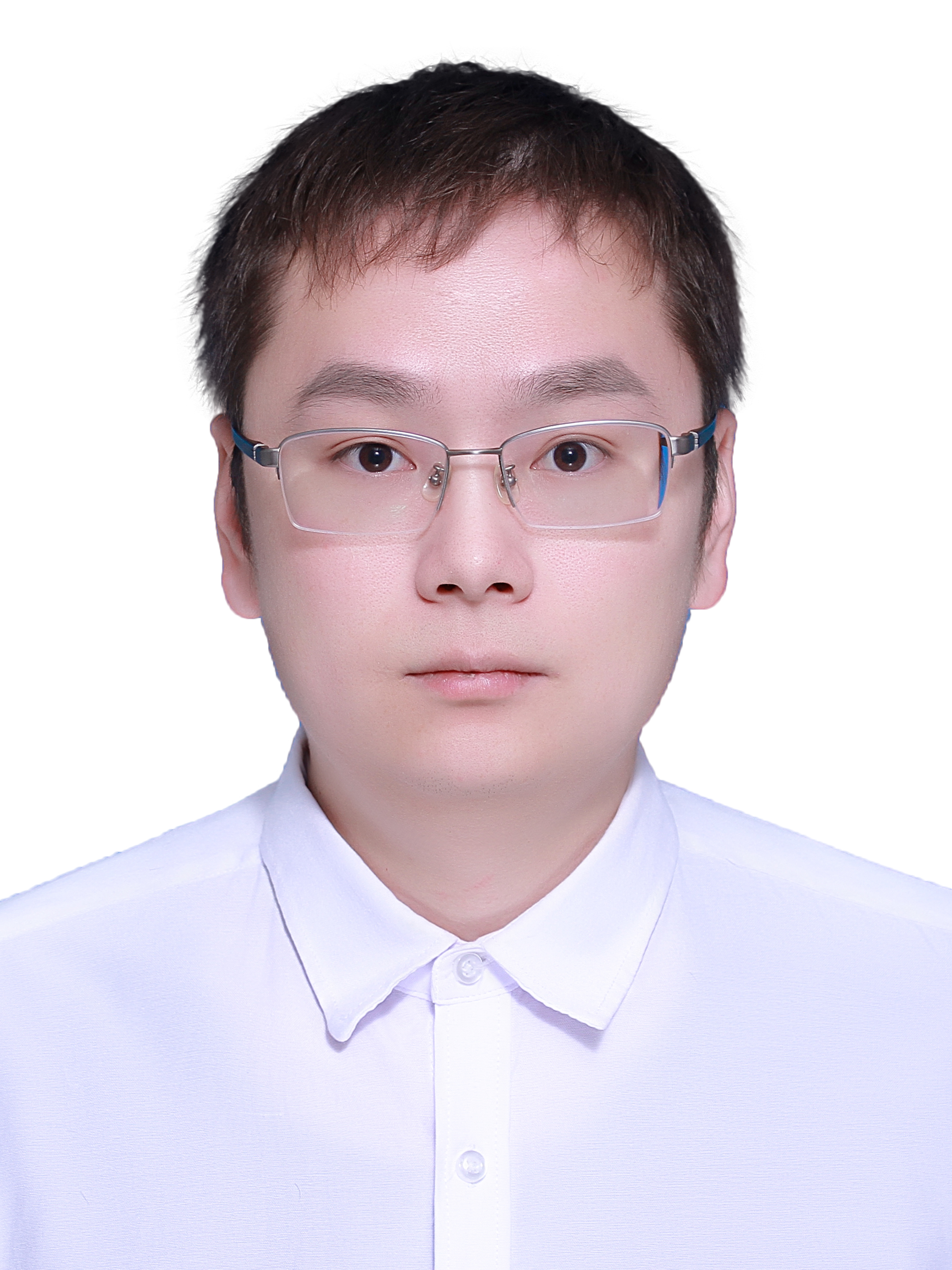}}]{Kun Zhang}
			received the PhD degree in computer science and technology from University of Science and Technology of China, Hefei, China, in 2019. He is is currently a faculty member with the Hefei University of Technology (HFUT), China. His research interests include Natural Language Understanding, Recommendation System. He has published several papers in refereed journals and conferences, such as the IEEE Transactions on Systems, Man, and Cybernetics: Systems, IEEE Transactions on Neural Networks and Learning Systems, the ACM Transactions on Knowledge Discovery from Data, AAAI, KDD, ACL, ICDM. He received the KDD 2018 Best Student Paper Award.
		\end{IEEEbiography}
		\vspace{-10mm}
		\begin{IEEEbiography}[{\includegraphics[width=1in,height=1.25in,clip,keepaspectratio]{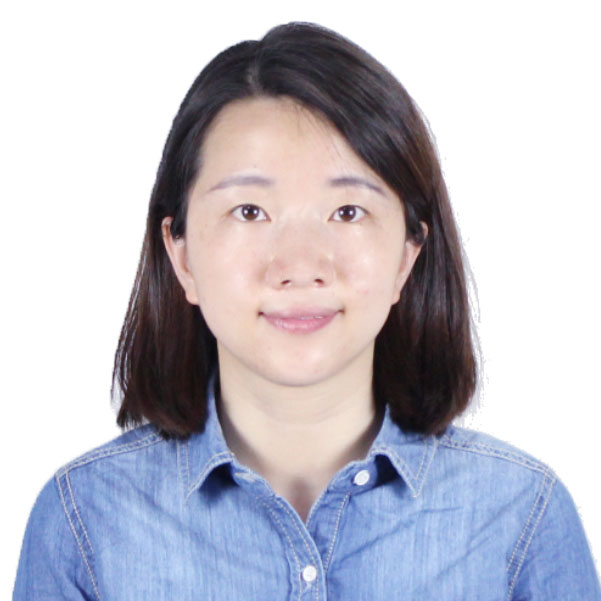}}]{Le Wu}
			received the PhD degree in computer science from the University of Science and Technology of China (USTC). She is a professor with Hefei University of Technology (HFUT), China. Her general area of research is data mining, recommender system, and social network analysis. She has published several papers in referred journals and conferences, such as the IEEE Transactions on Knowledge and Data Engineering, the ACM Transactions on Intelligent Systems and Technology, AAAI, IJCAI, KDD, SDM, and ICDM. She is the recipient of the Best of SDM 2015 Award.
		\end{IEEEbiography}
		\vspace{-10mm}
		\begin{IEEEbiography}[{\includegraphics[width=1in,height=1.25in,clip,keepaspectratio]{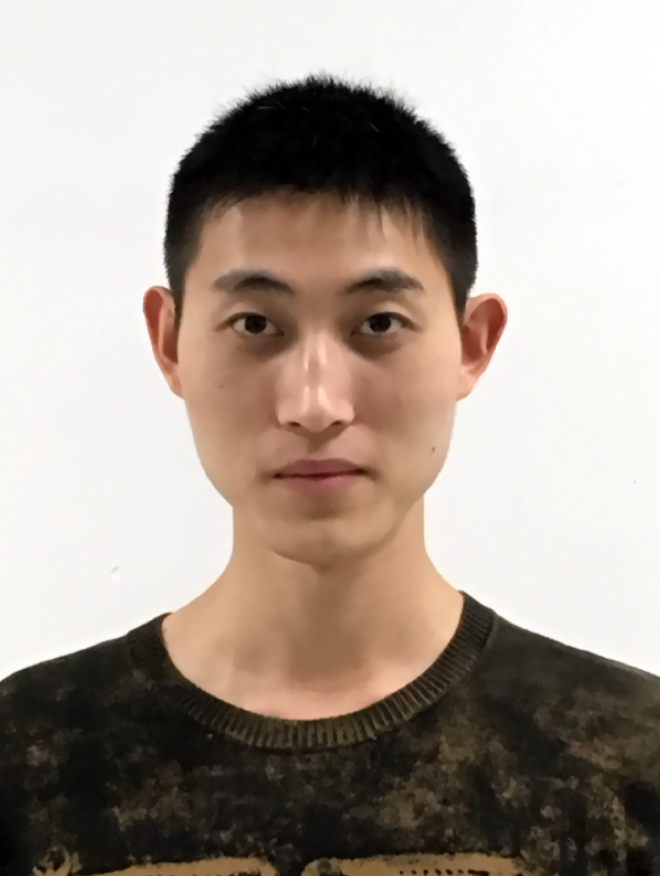}}]{Guangyi Lv}
			received the PhD degree in computer science and technology from University of Science and Technology of China, Hefei, China, in 2019. He is currently an Advisoary Researcher at Lenovo Research. His major research interests include deep learning, natural language processing and adaptive AI. He has published several papers in refereed journals and conferences, such as the IEEE Transactions on Systems, Man, and Cybernetics: Systems, the IEEE Transactions on Big Data, AAAI, IJCAI, ICDM. 
		\end{IEEEbiography}
		
		\vspace{-10mm}
		\begin{IEEEbiography}[{\includegraphics[width=1in,height=1.25in,clip,keepaspectratio]{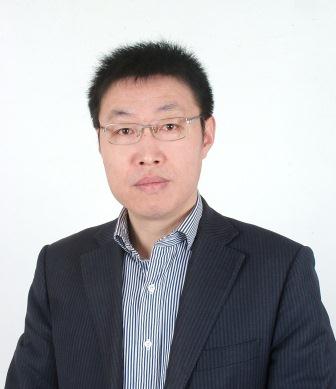}}]{Enhong Chen}
			(SM’07) received the PhD degree from USTC. He is a professor and vice dean of the School of Computer Science, USTC. His general area of research includes data mining and machine learning, social network analysis, and recommender systems. He has published more than 100 papers in refereed conferences and journals, including the IEEE Transactions on Knowledge and Data Engineering, the IEEE Transactions on Mobile Computing, KDD, ICDM, NIPS, and CIKM. He was on program committees of numerous conferences including KDD, ICDM, and SDM. His research is supported by the National Science Foundation for Distinguished Young Scholars of China. He is a senior member of the IEEE.
		\end{IEEEbiography}
		\vspace{-10mm}
		\begin{IEEEbiography}[{\includegraphics[width=1in,height=1.25in,clip,keepaspectratio]{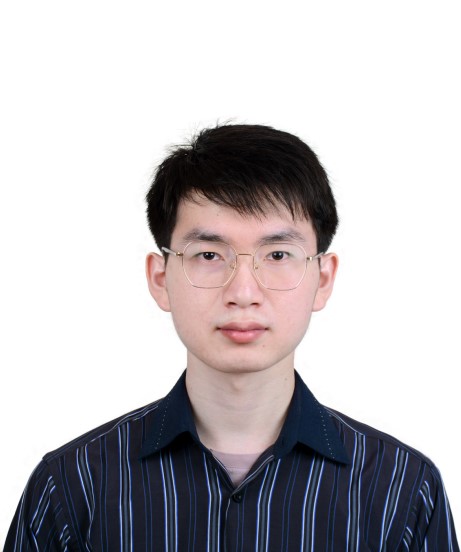}}]{ShuLan Ruan}
			received the B.S. degree from Hunan University, Changsha, China, in 2018. He is currently working toward the Ph.D. degree with the School of Computer Science and Technology, University of Science and Technology of China, Hefei, China. His research interests include sentiment analysis, computer vision and natural language processing. He has published several papers in AAAI and ICME.
		\end{IEEEbiography}
		\vspace{-10mm}
		\begin{IEEEbiography}[{\includegraphics[width=1in,height=1.25in,clip,keepaspectratio]{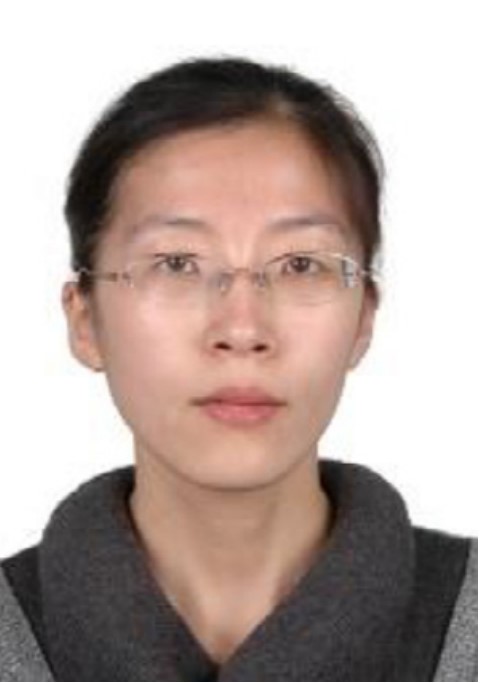}}]{Jing Liu}
			received the B.E. and M.S. degrees from Shandong University, Shandong, in 2001 and 2004, respectively, and the Ph.D. degree from the Institute of Automation, Chinese Academy of Sciences, Beijing, in 2008. She is a Professor with the National Laboratory of Pattern Recognition, Institute of Automation, Chinese Academy of Sciences. Her current research interests include deep learning, image content analysis and classification, multimedia, understanding and retrieval.
		\end{IEEEbiography}
		\vspace{-10mm}
		\begin{IEEEbiography}[{\includegraphics[width=1in,height=1.25in,clip,keepaspectratio]{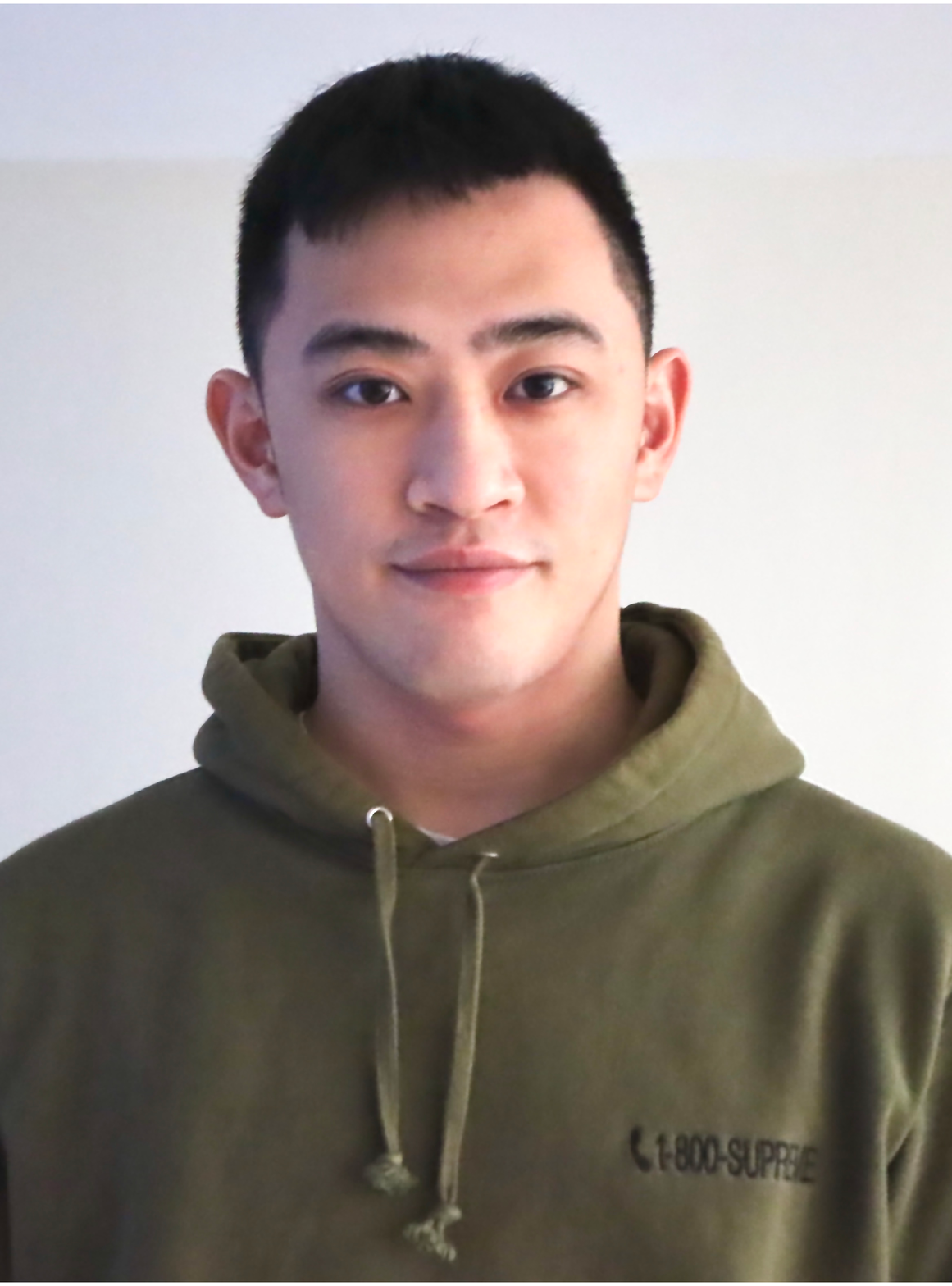}}]{Zhiqiang Zhang}
			is currently a Staff Engineer at Ant Group. His research interests mainly focus on graph machine learning. He has led a team to build an industrial graph machine learning system, AGL, in Ant Group. He has published more than 30 paper in top-tier machine learning and data mining conferences, including NeurIPS, VLDB, SIGKDD, and AAAI.
		\end{IEEEbiography}
		\vspace{-10mm}
		\begin{IEEEbiography}[{\includegraphics[width=1in,height=1.25in,clip,keepaspectratio]{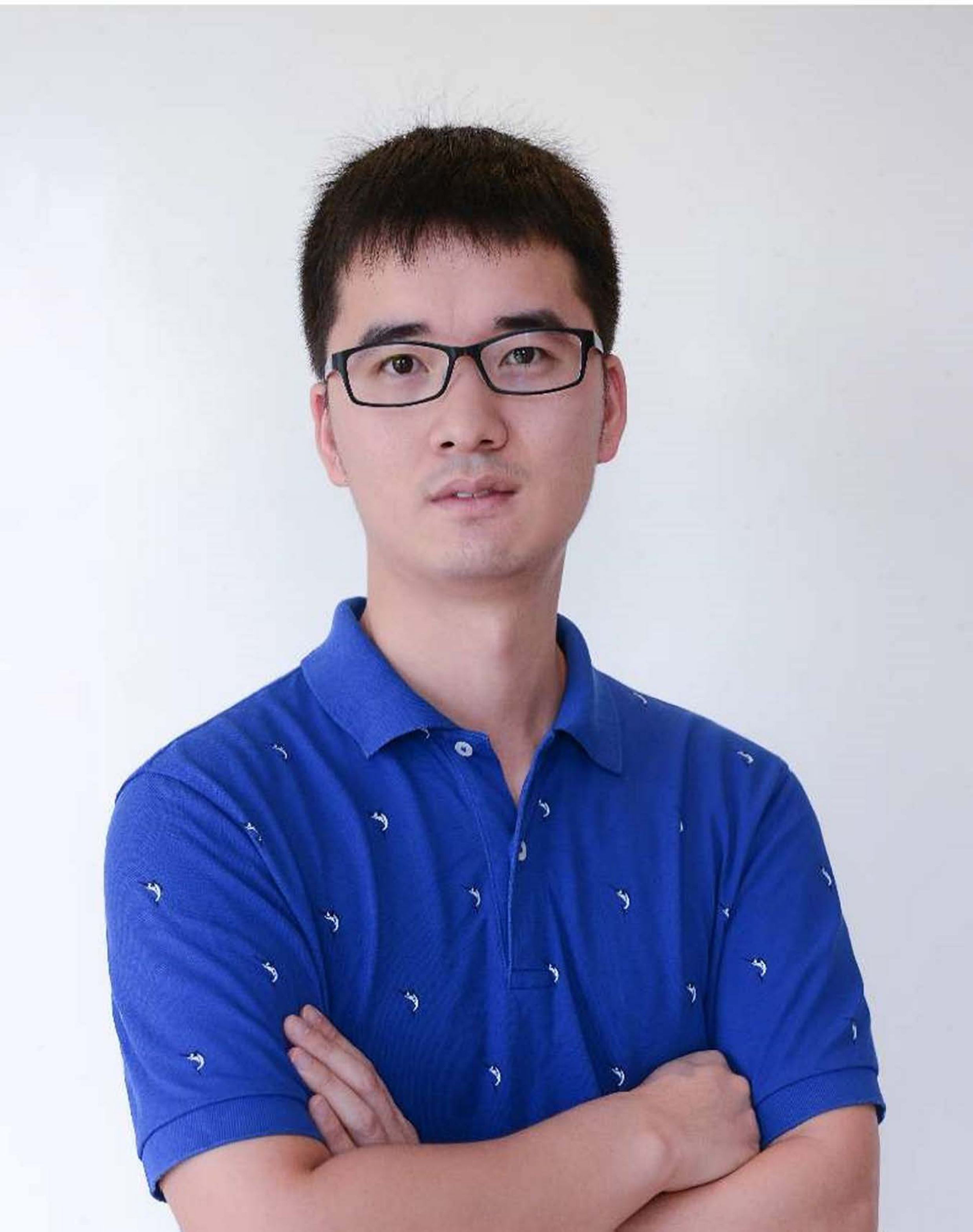}}]{Jun Zhou}
			is currently a Senior Staff Engineer at Ant Group. His research mainly focuses on machine learning and data mining. He has participated in the development of several distributed systems and machine learning platforms in Alibaba and Ant Group, such as Apsaras (Distributed Operating System), MaxCompute (Big Data Platform), and KunPeng (Parameter Server). He has published more than 40 papers in top-tier machine learning and data min-ing conferences, including VLDB, WWW, NeurIPS, and AAAI.
		\end{IEEEbiography}
		\vspace{-10mm}
		\begin{IEEEbiography}[{\includegraphics[width=1in,height=1.25in,clip,keepaspectratio]{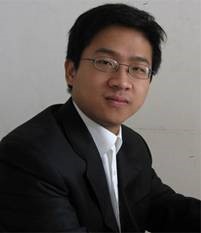}}]{Meng Wang}
			(SM’17) received the BE and PhD degrees from USTC, in 2003 and 2008, respectively. He is a professor with HFUT. His current research interests include multimedia content analysis, computer vision, and pattern recognition. He has authored more than 200 book chapters, journal, and conference papers in these areas. He is the recipient of the ACM SIGMM Rising Star Award 2014. He is an associate editor of the IEEE Transactions on Knowledge and Data Engineering, the IEEE Transactions on Circuits and Systems for Video Technology, and the IEEE Transactions on Neural Networks and Learning Systems. He is an IEEE Fellow.
		\end{IEEEbiography}

		
		

	\end{document}